%% file: main.tex
\pdfoutput=1
\documentclass[11pt]{article}

\usepackage{coling}

\usepackage{times}
\usepackage{latexsym}

\usepackage[T1]{fontenc}
\usepackage[utf8]{inputenc}

\usepackage{microtype}

\usepackage{inconsolata}

\usepackage{soul}
\usepackage{url}
\usepackage[utf8]{inputenc}
\usepackage{graphicx}
\usepackage{amsmath}
\usepackage{amsthm}
\usepackage{booktabs}
\usepackage{algorithm}

\usepackage{algorithmic}
\usepackage{multirow}
\usepackage{multicol}
\usepackage{bm}
\usepackage{underscore}
\usepackage{amsfonts}
\usepackage{xcolor}
\usepackage{color, colortbl}
\usepackage{makecell}
\usepackage{enumitem}

\usepackage{amssymb}
\usepackage{balance}
\usepackage{colortbl}
\usepackage{listings}
\usepackage{tabularx}
\usepackage{tikz}
\usepackage{rotating} % 
\usepackage[fixed]{fontawesome5}
\usepackage{nicematrix}
\usepackage[switch]{lineno}

\usepackage[edges]{forest}
\definecolor{hiddendraw}{RGB}{205, 44, 36}
\definecolor{hidden-blue}{RGB}{194,232,247}
\definecolor{hidden-orange}{RGB}{243,202,120}
\definecolor{hidden-yellow}{RGB}{242,244,193}
\definecolor{uclablue}{rgb}{0.15, 0.45, 0.68}

\usepackage{makecell}

\definecolor{mygray}{gray}{.9}
\definecolor{gray2}{gray}{.78}
\definecolor{gray3}{gray}{.7}
\definecolor{gray4}{gray}{.6}
\definecolor{gray5}{gray}{.5}
\newcommand{\rmnum}[1]{\romannumeral #1}
\newcommand{\Rmnum}[1]{\expandafter\@slowromancap\romannumeral #1@}

\newtheorem{defn}{\hspace{-1mm} Definition}

\newcommand{\ours}{\textsc{SnAg}}
\title{Noise-powered Multi-modal Knowledge Graph Representation Framework}

\author{
  Zhuo Chen$^{\heartsuit \clubsuit}$, 
  Yin Fang$^{\heartsuit \clubsuit}$, 
  Yichi Zhang$^{\heartsuit \clubsuit}$,
  Lingbing Guo$^{\heartsuit \clubsuit}$,
  Jiaoyan Chen$^\spadesuit$,\\
  \textbf{Jeff Z. Pan$^{\diamondsuit}$,
  Huajun Chen$^{\heartsuit \clubsuit}$,
  Wen Zhang$^{\heartsuit \clubsuit}$\thanks{\quad Corresponding Author.}}\\
  $^\heartsuit$Zhejiang University  \quad
    {$^\spadesuit$The University of Manchester}  \quad
    {$^\diamondsuit$The University of Edinburgh}  \\
  $^\clubsuit$ZJU-Ant Group Joint Lab of Knowledge Graph \\
 \texttt{ 
    \{zhuo.chen,fangyin,zhangyichi2022,lbguo,huajunsir,zhang.wen\}@zju.edu.cn
  }\\
\faGithub \textbf{\tt \url{https://github.com/zjukg/SNAG}}
}

\begin{document}
\maketitle
\begin{abstract}
The rise of Multi-modal Pre-training highlights the necessity for a unified Multi-Modal Knowledge Graph (MMKG) representation learning framework. Such a framework is essential for embedding structured knowledge into multi-modal Large Language Models effectively, alleviating issues like knowledge misconceptions and multi-modal hallucinations.
In this work, we explore the efficacy of models in accurately embedding entities within MMKGs through two pivotal tasks:  Multi-modal Knowledge Graph Completion (MKGC) and Multi-modal Entity Alignment (MMEA). 
Building on this foundation, 
we propose a novel \textbf{\ours}~method that utilizes a Transformer-based architecture equipped with modality-level noise masking to robustly integrate multi-modal entity features in KGs. By incorporating specific training objectives for both MKGC and MMEA, our approach achieves SOTA performance across a total of ten datasets, demonstrating its versatility.
Moreover, \textbf{\ours}~can not only function as a standalone model but also enhance other existing methods, providing stable performance improvements. Code and data are available at 
{{\color{blue} \url{https://github.com/zjukg/SNAG}}}.
\end{abstract}

% ---------------------------------------------------------
% --------------------  Introduction ----------------------
% ---------------------------------------------------------
\vspace{-3pt}
\section{Introduction}
% The exploration of multi-modal dimensions within Knowledge Graphs (KGs) has become a pivotal force in the semantic web domain, catalyzing advancements in various artificial intelligence applications. 
% With the evolution of Large language Models (LLMs) and Multi-modal Pre-training, the imperative for a robust and comprehensive Multi-Modal Knowledge Graph (MMKG) representation learning framework has become apparent. Such a framework is essential for the effective integration of structured knowledge into multi-modal LLMs at scale, addressing prevalent challenges like knowledge misconceptions and multi-modal hallucination.
Current efforts to integrate MMKG with pre-training are scarce. 
\textbf{Triple-level} methods~\cite{DBLP:conf/nips/PanYHSH22} treat triples as standalone knowledge units, embedding the \textit{(head entity, relationship, tail entity)} structure into Visual Language Model's space. On the other hand, \textbf{Graph-level} methods~\cite{DBLP:journals/corr/abs-2302-06891,DBLP:journals/corr/abs-2309-13625} capitalize on the structural connections among entities in a global MMKG. By selectively gathering multi-modal neighbor nodes around each entity featured in the training corpus, they apply techniques such as Graph Neural Networks (GNNs) or concatenation to effectively incorporate knowledge during the pre-training process.

However, these approaches predominantly view MMKG from a traditional KG perspective, not fully separating the MMKG representation process from downstream or pre-training tasks.
% In this work, we revisit MMKG representation learning from the unique perspective of MMKG itself, focusing on two principal tasks: Multi-modal Knowledge Graph Completion (MKGC) and Multi-modal Entity Alignment (MMEA), which serve as benchmarks for assessing multi-modal entity representation. We introduce a unified Transformer-based framework (\ours), infused with active noise injection, that not only refines MMKG representation but also demonstrates SOTA results across a diverse array of ten MKGC and MMEA datasets. 
% \ours~stands out for its lightweight design, efficiency, and adaptability, incorporating components like Entity-Level Modality Interaction that can be seamlessly upgraded with advanced technologies. A distinctive feature of our approach is the Gauss Modality Noise Masking module, which significantly boosts performance across various MKGC and MMEA methodologies. This module's design sharply contrasts with previous MMKG-related efforts, which primarily focus on designing methods to refuse and combat noise in MMKGs. In contrast, as shown in Figure \ref{fig:intro}, \ours~accepts and deliberately incorporates noise, adapting to the noisy real-world scenarios. Crucially, our strategy provides a theoretical foundation for unified entity representations, paving the way for future explorations in MMKG Pre-training and Multi-modal Knowledge Injection.
In this work, we revisit MMKG representation learning uniquely from the MMKG perspective itself, employing two tasks: Multi-modal Knowledge Graph Completion (MKGC) and Multi-modal Entity Alignment (MMEA) to validate our method. 

Specifically, we introduce a unified Transformer-based framework (\ours) that achieves SOTA results across an array of ten datasets by simply aligning it with task-specific Training targets.
\ours~stands out for its \textbf{parameter-efficient} design and \textbf{adaptability}, incorporating components like Entity-Level Modality Interaction that can be seamlessly upgraded with advanced technologies.
% A key aspect of our method is the Gauss Modality Noise Masking module, which can significantly boost performance across various MKGC and MMEA approaches.
% This module's design sharply contrasts with previous MMKG-related efforts, which primarily focus on designing methods to refuse and combat noise in MMKGs. In contrast, as shown in Figure \ref{fig:intro}, \ours~accepts and deliberately incorporates noise, adapting to the noisy real-world scenarios.
\begin{figure}[!tbp]
  \centering
     \vspace{-1pt}
   \vskip -0.2in
  \includegraphics[width=0.88\linewidth]{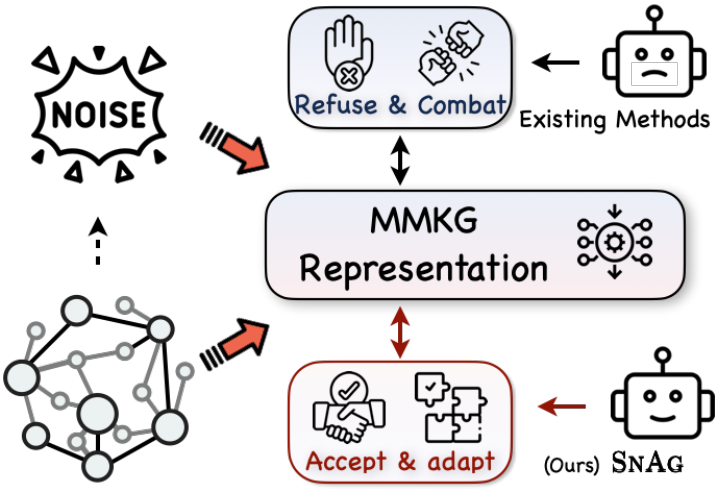}
  \vskip -0.1in
  \caption{Unlike existing models that refuse and combat noise in MMKGs, our \ours~accepts and deliberately incorporates noise to mirror noisy real-world scenarios. }
  \label{fig:intro}
   \vspace{-8pt}
   \vskip -0.15in
\end{figure}

A key aspect of our method is the Gauss Modality Noise Masking module, whose design sharply contrasts with previous MMKG-related efforts that primarily focus on designing methods to refuse and combat noise in MMKGs. In contrast, as shown in Fig.~\ref{fig:intro}, our \ours~accepts and deliberately incorporates noise, adapting to the noisy real-world scenarios. 
% This strategy can significantly boost performance across various MKGC and MMEA approaches.
Drawing inspiration from traditional mask-based multi-modal Pre-trained Language Models (PLMs) that enhance cross-modal alignment at the token level, our strategy innovates by \textbf{applying masking at the modality level}, significantly enhancing model's MMKG representation capabilities.
% Importantly, this work demonstrates the adaptability of our strategy for MKGC and MMEA, suggesting its potential to interface with more training tasks in the future and paving the way for further research in MMKG Pre-training and Multi-modal Knowledge Injection.
Importantly, as the first MMKG effort to concurrently support both MKGC and MMEA tasks, this work demonstrates its adaptability of our strategy, highlighting its potential to interface with more training tasks in the future and paving the way for further research in MMKG Pre-training and Multi-modal Knowledge Injection.

% ---------------------------------------------------------
% --------------------  Related Work ----------------------
% ---------------------------------------------------------
\vspace{-3pt}
\section{Related Work}\label{sec:pre}
% Typically, a KG is considered multi-modal when it contains knowledge symbols expressed across various modalities, including, but not limited to, text, images, sound, or video~\cite{DBLP:journals/corr/abs-2402-05391}. 
% Current research primarily concentrates on the visual modality, assuming that other modalities can be processed similarly.

\vspace{-2pt}
\subsection{MMKG Representation}
The current mainstream approaches to MMKG representation learning can broadly be classified into two distinct categories:
\textbf{\textit{(\rmnum{1})}}~\textbf{Late Fusion} methods emphasize modality interactions and feature aggregation just prior to output generation.
For example,
MKGRL-MS~\cite{DBLP:journals/inffus/WangYCSL22}  crafts unique single-modal embeddings, employing multi-head self-attention to determine each modality's contribution to semantic composition and \textbf{sum} the weighted multi-modal features for MMKG entity representation. 
MMKRL~\cite{DBLP:journals/apin/LuWJHL22-MMKRL} learns cross-modal embeddings in a unified translational semantic space, merging them through \textbf{concatenation}.
DuMF~\cite{DBLP:journals/asc/LiZWZH22} applies a bilinear layer for feature projection and an attention block for modality preference learning in each track, integrating features via a \textbf{gate network}.
\textbf{\textit{(\rmnum{2})}}~\textbf{Early Fusion}
methods integrate multi-modal feature at an initial stage, enabling full modality interactions for complex reasoning.
For example, \citet{DBLP:journals/tkde/FangZHWX23} first normalizes entity modalities into a unified embedding using an MLP, then refines them by contrasting with perturbed negative samples.
MMRotatH~\cite{DBLP:journals/kais/WeiCWLZ23} utilizes a gated encoder to merge textual and structural data, filtering irrelevant information within a rotational dynamics-based KGE framework.
Recent studies \cite{DBLP:conf/sigir/ChenZLDTXHSC22-MKGformer,DBLP:conf/emnlp/LeeCLJW23-VISITA}  utilize (V)PLMs like BERT and ViT for multi-modal data integration.  These methods convert graph structures, text, and images into sequences or dense embeddings suited for LMs, leveraging the LMs' reasoning abilities and embedded knowledge for tasks like Multi-modal Link Prediction.  However, they rely heavily on pre-trained models, resulting in significant parameter sizes and training costs~\cite{DBLP:journals/corr/abs-2402-05391}.

In this paper, we propose a Transformer-based method \textbf{\ours}~that introduces fine-grained, entity-level modality {preference} to enhance entity representation. This strategy combines the benefits of Early Fusion, with its effective modality interaction, while also aligning with the Late Fusion modality integration paradigm. Furthermore, our model is lightweight, with \textbf{only 13M parameters}, far fewer than traditional PLM-based methods, which often exceed \textbf{200M parameters}. This offers increased flexibility and wider applicability.

\begin{figure*}[!tbp]
  \centering
  \vskip -0.2in
  \includegraphics[width=1.0\linewidth]{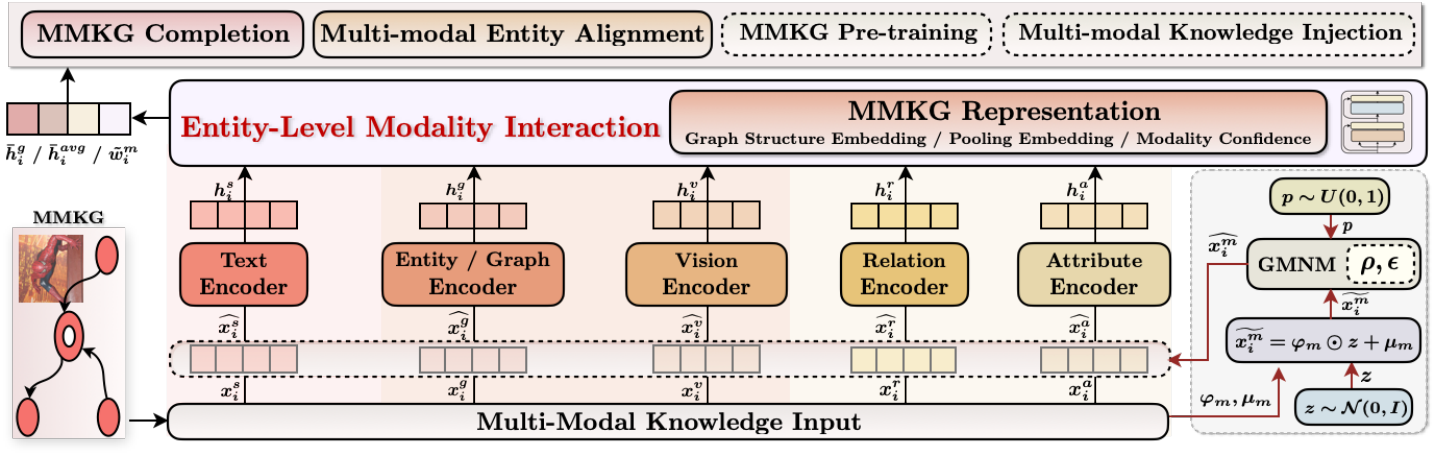}
  \vspace{-15pt}
  \caption{The overall framework of \ours. }
  \label{fig:model}
   \vspace{-5pt}
   \vskip -0.15in
\end{figure*}

\subsection{Multi-Modal Knowledge Graph Completion and Alignment}
Multi-modal Knowledge Graph Completion (MKGC) is crucial for inferring missing triples in existing MMKGs~\cite{DBLP:conf/emnlp/LeeCLJW23-VISITA,DBLP:conf/emnlp/ZhaoCWZZZJ22-MOSE}.
Entity Alignment (EA) focuses on KG integration, aiming to identify identical entities across different KGs by leveraging relational, attributive, and literal (surface) features. Multi-Modal Entity Alignment (MMEA) enhances this by incorporating visual data, thereby improving alignment accuracy~\cite{DBLP:conf/ksem/ChenLWXWC20,DBLP:journals/corr/abs-2310-06365}. Further details are provided in Appendix~\ref{sec:relat-sup}.
Despite nearly five years of development, MMEA and MKGC have progressed independently within the MMKG field, lacking a unified framework that integrates these tasks. Given the advancements in multi-modal LLMs, it is timely to develop a comprehensive framework that addresses both MKGC and MMEA, offering enhanced multi-modal entity representations.
% ---------------------------------------------------------
% -----------------------  Method -------------------------
% ---------------------------------------------------------
% \vspace{-3pt}
\section{Method}\label{sec:mtd}
\subsection{Preliminaries}
% Drawing on the categorization proposed in~\cite{DBLP:journals/corr/abs-2202-05786}, we distinguish between two types of MMKGs: A-MMKG and N-MMKG. In A-MMKGs, images are attached to entities as attributes, while in N-MMKGs, images are treated as standalone entities interconnected with others. A-MMKGs are more prevalent in current research and applications within the semantic web community due to their accessibility and similarity to traditional KGs~\cite{DBLP:journals/corr/abs-2402-05391}.  Therefore, this paper will focus exclusively on A-MMKG, unless stated otherwise.
This paper focuses on A-MMKG~\cite{DBLP:journals/corr/abs-2202-05786}, where images are attached to entities as attributes.

\begin{defn}{\textbf{Multi-modal Knowledge Graph.}}\label{def:mmkg}
A KG is defined as $\mathcal{G}=\{\mathcal{E}, \mathcal{R}, \mathcal{A}, \mathcal{T}, \mathcal{V}\}$ where $\mathcal{T} = \{\mathcal{T_A}, \mathcal{T_R}\}$ with $\mathcal{T_R}=\mathcal{E}\times\mathcal{R}\times\mathcal{E}$ and $\mathcal{T_A}=\mathcal{E}\times\mathcal{A}\times\mathcal{V}$. 
MMKG utilizes multi-modal data (e.g., images) as specific attribute values for entities or concepts, with $\mathcal{T_A}=\mathcal{E}\times\mathcal{A}\times(\mathcal{V}_{KG}\cup\mathcal{V}_{MM})$, where $\mathcal{V}_{MM}$ are values of multi-modal data (e.g., images).
% For instance, in an MMKG, an attribute triple $(e, a, v)$ in $\mathcal{T_A}$ might associates an image as $v$ to an entity $e$ via an attribute $a$, typically denoted as \textit{hasImage}.
\end{defn}

\begin{defn}{\textbf{MMKG Completion.}}\label{def:mmkgc}
The objective is to augment the set of relational triples $\mathcal{T}_{R}$ within MMKGs by identifying and adding missing relational triples among existing entities and relations, potentially 
 utilizing attribute triples $\mathcal{T_A}$. Specifically, our focus is on Entity Prediction, which involves determining the missing head or tail entities in queries of the form $(head, r, ?)$ or $(?, r, tail)$.
\end{defn}

\begin{defn}{\textbf{Multi-modal Entity Alignment.}}\label{def:mmea}
Given two aligned {MMKGs} $\mathcal{G}_1$ and $\mathcal{G}_2$, the objective of MMEA is to identify entity pairs ($e^1_i$, $e^2_i$) from $\mathcal{E}_1$ and $\mathcal{E}_2$, respectively, that correspond to the same real-world entity $e_i$. This process utilizes a set of pre-aligned entity pairs, divided into a training set (seed alignments $\mathcal{S}$) and a testing set $\mathcal{S}_{te}$, following a pre-defined seed alignment ratio $R_{sa}$ $=$ $|\mathcal{S}| / |\mathcal{S}\cup\mathcal{S}_{te}|$.
The modalities associated with an entity are denoted by $\mathcal{M}=\{{g}, {r}, {a}, {v}, {s}\}$, signifying graph structure, relation, attribute, vision, and surface (i.e., entity names) modalities, respectively.
\end{defn}

\vspace{-4pt}
\subsection{Multi-Modal Knowledge Embedding} \label{def:mmke}

% \subsubsection{\textbf{Graph Structure Embedding}} 
\paragraph{Graph Structure Embedding.}
Let $x_i^g \in \mathbb{R}^d$ represents the graph embedding of entity $e_i$, which is randomly initialized and learnable, with $d$ representing the predetermined hidden dimension.
In \ul{MKGC}, we follow \citet{zhang2024unleashing} to set $h_i^g$ $=$ $FC_g({W}_g, x_i^g)$, where $FC_g$ is a KG-specific fully connected layer applied to $x_i^g$ with weights ${W}_g$.
For \ul{MMEA}, we follow~\citet{chen2023meaformer} to utilize the Graph Attention Network (GAT)~\cite{DBLP:conf/iclr/VelickovicCCRLB18}, configured with two attention heads and two layers, to capture the structural information of $\mathcal{G}$.
This is facilitated by a diagonal weight matrix~\cite{DBLP:journals/corr/YangYHGD14a} ${W}_g \in \mathbb{R}^{d \times d}$ for linear transformation.
The structure embedding is thus defined as $h_i^g$ $=$ $GAT({W}_g, {M}_g; x_i^g)$, where ${M}_g$ refers to the graph's adjacency matrix.

\vspace{-2pt}
% \subsubsection{\textbf{Relation and Attribute Embedding}} 
\paragraph{Relation and Attribute Embedding.}
Our \ul{MKGC} study aligns with domain practices~\cite{DBLP:conf/emnlp/ZhaoCWZZZJ22-MOSE,DBLP:conf/www/LiZXZX23-IMF} which focuses exclusively on relation triples. These are represented by learnable embeddings $x_j^r\,\in\,\mathbb{R}^{d/2}$, where $j$ uniquely identifies each relation $r_j$, distinguishing it from entity indices. We exclude attribute triples to maintain consistency with methodological practices in the field. The choice of setting a dimensionality of $d/2$ is based on our application of the RotatE model~\cite{DBLP:conf/iclr/SunDNT19}, which assesses triple plausibility.
RotatE interprets relations as rotations in a complex space, requiring the relation embedding's dimension to be half that of the entity embedding to account for the real and imaginary components of complex numbers.
For \ul{MMEA}, following \citet{DBLP:conf/emnlp/YangZSLLS19}, we use bag-of-words features for relation ($x^r$) and attribute ($x^a$) representations of entities (detailed in \S\,\ref{sec:detail}) . Separate FC layers, parameterized by ${W}_m \in \mathbb{R}^{d_m \times d}$, are employed for embedding space harmonization: $h_i^m$ $=$ $FC_m({W}_m, x_i^m)$, where $m \in \{r, a\}$ and $x_i^m \in \mathbb{R}^{d_m}$ represents the input feature of entity $e_i$ for modality $m$.

\vspace{-2pt}
% \subsubsection{\textbf{Visual and Surface Embedding}} \label{sec:img}
\paragraph{Visual and Surface Embedding.}\label{sec:img}
For visual embeddings, a pre-trained (and thereafter frozen) visual encoder, denoted as $Enc_v$, is used to extract visual features $x^v_i$ for each entity $e_i$ with associated image data. In cases where entities lack corresponding image data, we synthesize random image features adhering to a normal distribution, parameterized by the mean and standard deviation observed across other entities' images~\cite{chen2023meaformer,chen2023rethinking,zhang2024unleashing}.
Regarding surface embeddings for MKGC, we leverage Sentence-BERT~\cite{DBLP:conf/emnlp/SBERT}, a pre-trained textual encoder, to derive textual features from each entity's description. The {\tt [CLS]} token serves to aggregate sentence-level textual features $x^s_i$.
Consistent with the approach applied to other modalities, we utilize 
% separate fully connected layers, denoted as 
$FC_m$ parameterized by ${W}_m \in \mathbb{R}^{d_m \times d}$ to integrate the extracted features $x_i^v$ and $x_i^s$ into the embedding space, yielding the embeddings $h_i^v$ and $h_i^s$.

\vspace{-2pt}
\subsection{Gauss Modality Noise Masking}\label{sec:noise}
\vspace{-1pt}
Recent research in MMKG \cite{chen2023rethinking,DBLP:journals/corr/abs-2305-14651} suggests that models  can tolerate certain noise levels without  a noticeable decline in the expressive capability of multi-modal entity representations, a finding echoed across various machine learning domains~\cite{DBLP:journals/corr/abs-2310-05914,chen2024learning}.
Additionally, 
% \citet{DBLP:journals/corr/abs-2401-14887} observe that in the Retrieval-Augmented Generation (RAG) process of LLMs, filling up the retrieved context with irrelevant documents consistently improves model performance in realistic scenarios.
% Similarly, 
\citet{DBLP:conf/aaai/ChenHCGZFPC23} demonstrate that cross-modal masking and reconstruction can improve a model's cross-modal alignment capabilities in Zero-shot Image Classification scenario.
Inspired by evidence of model noise resilience, we hypothesize that introducing noise during MMKG modality fusion training could enhance both modal feature robustness and real-world performance.
In light of these observations, 
we propose a new mechanism termed Gauss Modality Noise Masking (GMNM),  aimed at enhancing modality feature representations through controlled noise injection at the training stage for MMKG. This stochastic strategy introduces a probabilistic transformation to each modality feature $x_i^m$ at the beginning of every training epoch, described as :
\begin{equation}
\vspace{-1pt}
\widehat{x_i^m} =
\begin{cases}
x_i^m, & \text{if } p > \rho\,, \\
(1-\epsilon)x_i^m + \epsilon \widetilde{x_i^m}\,, & \text{otherwise},
\end{cases}
\label{eq:1}
\vspace{-1pt}
\end{equation}
where $p$ $\sim$ $U(0, 1)$ denotes a uniformly distributed random variable 
% dictating the application of noise, and $\rho$ represents the probability of applying this noise for each $x_i^m$. $\epsilon$ is the noise ratio. 
 that determines whether noise is applied, with $\rho$ being the threshold probability for noise application to each $x_i^m$. Here, $\epsilon$ signifies the noise (mask) ratio.
We define the generation of noise vector $\widetilde{x_i^m}$ as:
\begin{equation}
\vspace{-1pt}
\widetilde{x_i^m} = \varphi_m \odot z + \mu_m\,\,, \quad z \sim \mathcal{N}({0},{I}),
\vspace{-1pt}
\end{equation}
where $\varphi_m$ and $\mu_m$ represent the standard deviation and mean of the \textbf{modality-specific non-noisy data} for $m$, respectively, and $z$ denotes a sample drawn from a Gaussian distribution $\mathcal{N}(0, I)$ with mean vector with mean 0 and identity covariance matrix $I$, ensuring that the introduced noise is statistically coherent with the intrinsic data variability of the respective modality.
Additionally, the intensity of noise ($\epsilon$) can be dynamically adjusted to simulate real-world data imperfections. This adaptive noise injection strategy is designed to foster a model resilient to data variability, capable of capturing and representing complex multi-modal interactions with enhanced fidelity in practical applications.

Note that after the transformation from $x^m$ to $\widehat{x^m}$, these modified features are still subject to further processing through $FC_m$ as detailed in \S\,\ref{def:mmke}. This critical step secures the generation of the ultimate modal representation, symbolized as $\widehat{h^m}$. 
For clarity in subsequent sections, \textbf{we will treat $h^m$ and $h^m_i$ as representing their final states, $\widehat{h^m}$ and $\widehat{h^m_i}$}, unless specified otherwise.

\vspace{-2pt}
\subsection{Entity-Level Modality Interaction}\label{sec:inter}
This phase is designed for instance-level modality weighting and fusion, enabling dynamic adjustment of training weights  based on modality information's signal strength and noise-induced uncertainty. We utilize a Transformer architecture~\cite{DBLP:conf/nips/VaswaniSPUJGKP17} for this purpose, noted for its efficacy in modality fusion and its ability to derive confidence-based weighting for modalitieswhich improves interpretability and adaptability.  
% The Transformer's self-attention mechanism is crucial for evaluating modal input reliability, ensuring the model prioritizes accurate and relevant information.
The Transformer's self-attention mechanism is crucial for ensuring the model evaluates and prioritizes accurate and relevant modal inputs.

Specifically, we adapt the vanilla Transformer  through integrating three key components: Multi-Head Cross-Modal Attention (MHCA), Fully Connected Feed-Forward Networks (FFN), and Instance-level Confidence (ILC).

\noindent \textbf{\textit{(\rmnum{1})} MHCA} operates its attention function across $N_h$ parallel heads. Each head, indexed by $i$, employs shared matrices ${W}_q^{(i)}$, ${W}_k^{(i)}$, ${W}_v^{(i)}$ $\in \mathbb{R}^{d \times d_h}$ (where $d_h=d/N_h$), to transform input $h^m$ into queries ${Q}^{(i)}_m$, keys ${K}^{(i)}_m$, and values ${V}^{(i)}_m$:
\begin{equation*}
{Q}^{(i)}_m, {K}^{(i)}_m, {V}^{(i)}_m =h^m {W}_q^{(i)}, h^m {W}_k^{(i)}, h^m {W}_v^{(i)} \,.\\
\end{equation*}
The output for modality $m$'s feature is then generated by combining the outputs from all heads and applying a linear transformation:
\begin{align}
{MHCA}(h^m) & =\bigoplus\nolimits_{i=1}^{N_h}{head}_i^m \cdot {W}_{0} \,, \label{eq:4} \\
{head}_i^m & = \sum\nolimits_{j \in \mathcal{M}} \beta^{(i)}_{mj}{V}^{(i)}_j \,, \label{eq:5}
\end{align}
where ${W}_{0}$ $\in \mathbb{R}^{d \times d}$.  The attention weight $\beta_{mj}$ calculates the relevance between modalities $m$, $j$:
\begin{equation}
    \beta_{m j}=\frac{\exp (Q_m^\top K_j / \sqrt{d_h} )}{\sum_{i \in \mathcal{M}} \exp (Q_m^\top K_i / \sqrt{d_h})}\,.
\end{equation}
Besides, layer normalization (LN) and residual connection (RC) are incorporated to stabilize training:
\begin{equation}
\label{eq:hbar}
\bar{h}^m = LayerNorm({MHCA}(h^m) + h^m) \,.
\end{equation}

\noindent \textbf{\textit{(\rmnum{2})} FFN}: This network, consisting of two linear transformations and a ReLU activation, further processes the MHCA output:
\begin{align*}
{FFN}(\bar{h}^m) & = ReLU(\bar{h}^m{W}_{1} + b_{1}){W}_{2} +b_{2} \,, \\
\label{eq:hidden}
\bar{h}^m & \gets LayerNorm({FFN}(\bar{h}^m) + \bar{h}^m) \,,
\end{align*}
where ${W}_{1}$ $\in \mathbb{R}^{d \times d_{in}}$ and ${W}_{2}$ $\in \mathbb{R}^{d_{in} \times d}$.

\noindent \textbf{\textit{(\rmnum{3})} ILC}: To capture crucial inter-modal interactions and tailors the model’s confidence for each entity's modality, we calculate the confidence $\tilde{w}^m$:
\begin{equation}
\vspace{-0.5pt}
\scalebox{1.0}{$
    \tilde{w}^m = \frac{\exp(\sum\nolimits_{j \in \mathcal{M}} \sum\nolimits_{i=0}^{N_h}  \beta^{(i)}_{mj}/\sqrt{|\mathcal{M}| \times N_h})}{\sum\nolimits_{k \in \mathcal{M}}\exp(\sum\nolimits_{j \in \mathcal{M}} \sum\nolimits_{i=0}^{N_h}  \beta^{(i)}_{kj}\sqrt{|\mathcal{M}| \times N_h})}\,.
    $}
    \label{eq:10}
\vspace{-0.5pt}
\end{equation}

\vspace{-2pt}
\subsection{Task-Specific Training}
Building upon the foundational processes detailed in previous sections, we have derived multi-modal KG representations denoted as ${h}^m$ (discussed in \S\,\ref{sec:noise}) and $\bar{h}^m$ (elaborated in \S\,\ref{sec:inter}, Eq.~\eqref{eq:hbar}), along with confidence scores $\tilde{w}^m$ for each modality $m$ within the MMKG (introduced in \S\,\ref{sec:inter}, Eq.~\eqref{eq:10}).

% \subsubsection{\textbf{MMKG Completion}}
\paragraph{MMKG Completion.}
Within MKGC, we consider two methods for entity representation as candidates:
\textbf{\textit{(\rmnum{1})}} $\bar{h}^g$: Reflecting insights from previous research \cite{chen2023meaformer,zhang2024unleashing}, graph structure embedding emerges as crucial for model performance. After being processed by the Transformer layer, $\bar{h}^g$ not only maintains its structural essence but also blends in other modal insights (refer to Eq.~\eqref{eq:4} and \eqref{eq:5}), offering a comprehensive multi-modal entity representation.
\textbf{\textit{(\rmnum{2})}} $\bar{h}^{avg}$: For an equitable multi-modal representation, we average all modality-specific representations via
% \begin{equation}
$\bar{h}^{avg} = \frac{1}{|\mathcal{M}|} \sum\nolimits_{m \in \mathcal{M}} \bar{h}^m$,
% \end{equation}
where $\mathcal{M}$ is the set of all modalities. This averaging ensures equal modality contribution, leveraging the rich, diverse information within MMKGs.
For consistency in the following descriptions, we will refer to both entity representations using the notation $\bar{h}$.

We apply the RotatE model~\cite{DBLP:conf/iclr/SunDNT19} as our score function to assess the plausibility of triples. It is defined as:
\begin{equation}
\vspace{-0.5pt}
    \mathcal{F}(e^h, r,e^t)=||\bar{h}^{head} \circ x^{r} - \bar{h}^{tail}||\,,
\vspace{-0.5pt}
\end{equation}
where $\circ$ represents the rotation operation in complex space, which transforms the head entity's embedding by the relation to approximate the tail entity's embedding.

To prioritize positive triples with higher scores, we optimize the embeddings using a sigmoid-based loss function~\cite{DBLP:conf/iclr/SunDNT19}. The loss function is given by:
\begin{equation*}
\vspace{-0.5pt}
\scalebox{0.9}{$
    \begin{aligned}
    \mathcal{L}_{kgc}&=\frac{1}{|\mathcal{T_R}|}\sum_{(e^h, r,e^t)\in \mathcal{T_R}}\Big(-\log\sigma(\lambda-\mathcal{F}(e^h, r,e^t))\\
    &-\sum\nolimits_{i=1}^{K}\upsilon_i\log\sigma(\mathcal{F}(e^{h\prime}, r^\prime,e^{t\prime})-\lambda)\Big)\,,
    \end{aligned}
    $}
    \vspace{-0.5pt}
\end{equation*}
where $\sigma$ denotes the sigmoid function, $\lambda$ is the margin, $K$ is the number of negative samples per positive triple, and $\upsilon_i$ represents the self-adversarial weight for each negatively sampled triple $(e^{h\prime}, r^\prime, e^{t\prime})$. Concretely, $\upsilon_i$ is calculated as:
\begin{equation}
\vspace{-0.5pt}
    \upsilon_i=\frac{\exp(\tau_{kgc}\mathcal{F}(e^{h\prime}_i, r^\prime_i,e^{t\prime}_i))}{\sum_{j=1}^K \exp(\tau_{kgc}\mathcal{F}(e^{h\prime}_j, r^\prime_j,e^{t\prime}_j))}\,,
    \vspace{-0.5pt}
\end{equation}
with  $\tau_{kgc}$ being the temperature parameter. Our primary objective is to minimize  $\mathcal{L}_{kgc}$, thereby refining the embeddings to accurately capture MMKG's underlying relationships.

% \subsubsection{\textbf{Multi-modal Entity Alignment}}
\paragraph{Multi-modal Entity Alignment.}
In MMEA, following \cite{chen2023rethinking,chen2023meaformer}, we adopt the Global Modality Integration (GMI) derived multi-modal features as the representations for entities. GMI emphasizes global alignment by concatenating and aligning multi-modal embeddings with a learnable global weight, enabling adaptive learning of each modality's quality across two MMKGs.
The GMI joint embedding ${h}^{GMI}_i$ for entity $e_i$ is defined as:
\begin{align} \label{eq:cat}
\vspace{-0.5pt}
    {h}^{GMI}_i = \bigoplus\nolimits_{m \in \mathcal{M}}[w_m{h}_i^m]\, ,
\vspace{-0.5pt}
\end{align} 
where $\bigoplus$ signifies vector concatenation and $w_m$ is the global weight for modality $m$, which is distinct from the entity-level dynamic modality weights $\tilde{w}^m$ in Eq.~\eqref{eq:10}.

\input{tab/mkgc-tb}

We note that
the distinction between MMEA and MKGC lies in their focus: MMEA emphasizes aligning modal features between entities and distinguishing non-aligned entities, prioritizing original feature retention. In contrast, MKGC emphasizes the inferential benefits of modality fusion across different multi-modal entities. As demonstrated by \citet{chen2023rethinking}, the modality feature is often smoothed by the Transformer Layer in MMEA, potentially reducing entity distinction. GMI addresses this by preserving essential information, aiding alignment stability.

Moreover, as a unified MMKG representation framework, modal features extracted earlier are optimized through MMEA-specific training objectives~\cite{DBLP:conf/coling/LinZWSW022}. Specifically, for each aligned entity pair ($e_i^1$,$e_i^2$) in training set (seed alignments $\mathcal{S}$), we define a negative entity set $\mathcal{N}^{ng}_i$ $=$ $\{e^1_j|\forall e^1_j \in \mathcal{E}_1, j \neq i\}$ $\cup$ $\{e^2_j|\forall e^2_j \in \mathcal{E}_2, j \neq i\}$ and utilize in-batch ($\mathcal{B}$) negative sampling~\cite{DBLP:conf/icml/ChenK0H20} to enhance efficiency.  
The alignment probability distribution is:
\begin{equation*}\label{eq:macl}
    p_m(e^1_i, e^2_i) =  \frac{\gamma_m(e^1_i, e^2_i)}{\gamma_m(e^1_i, e^2_i) + \sum\nolimits_{e_j \in \mathcal{N}^{ng}_i}\gamma_m(e^1_i, e_j)} \, ,
\end{equation*}
where $\gamma_m(e_i, e_j)$ $=$ $\exp({h^{m\top}_{i}}{h^m_{j}}/\tau_{ea})$ and $\tau_{ea}$  is the temperature hyper-parameter. 
We establish a bi-directional alignment objective to account for MMEA directions:
\begin{equation*}\label{eq:clloss}
    \mathcal{L}_m = - \mathbb{E}_{i \in \mathcal{B}}\, \log[\,p_m(e^1_i, e^2_i)+p_m(e^2_i, e^1_i)\,]/2 \,,
\end{equation*}
\textbf{\textit{(\rmnum{1})}} The training objective denoted as $\mathcal{L}_{GMI}$ when using GMI joint embeddings, i.e., $\gamma_{GMI}(e_i, e_j)$ is set to $\exp({h^{{GMI}\top}_{i}}{h^{GMI}_{j}}/\tau_{ea})$.

To integrate dynamic confidences into the training process and enhance multi-modal entity alignment, we adopt two specialized training objectives from \citet{chen2023rethinking}: 
\textbf{\textit{(\rmnum{2})}} {Explicit Confidence-augmented Intra-modal Alignment (ECIA):} This objective modifies Eq.~\eqref{eq:clloss} to incorporate explicit confidence levels within the same modality, defined as: $\mathcal{L}_{ECIA} = \sum\nolimits_{m \in \mathcal{M}}\widetilde{\mathcal{L}}_m\,$, where:
\begin{equation*}
\vspace{-0.5pt}
\scalebox{0.82}{$
    \widetilde{\mathcal{L}}_m = - \mathbb{E}_{i \in \mathcal{B}}\,\log[\,\phi_m(e^1_i, e^2_i )*(p_m(e^1_i, e^2_i)+p_m(e^2_i, e^1_i))\,]/2 \,.
$}
\vspace{-0.5pt}
\end{equation*}
Here, $\phi_m(e^1_i, e^2_i)$ represents the minimum confidence value between entities $e^1_i$ and $e^2_i$ in modality $m$, i.e., $\phi_m(e_i, e_j) = Min(\tilde{w}_i^m, \tilde{w}_j^m)$, addressing the issue of aligning high-quality features with potentially lower-quality ones or noise.
\textbf{\textit{(\rmnum{3})}} {Implicit Inter-modal Refinement (IIR)} refines entity-level modality alignment by leveraging the transformer layer outputs $\bar{h}^m$, aiming to align output hidden states directly and adjust attention scores adaptively. The corresponding loss function is:
$\mathcal{L}_{IIR} = \sum\nolimits_{m \in \mathcal{M}}\bar{\mathcal{L}}_m\,,$
where $\bar{\mathcal{L}}_m$ is also a variant of $\mathcal{L}_m$ (Eq.~\eqref{eq:clloss}) with 
$\bar{\gamma}_m(e_i, e_j) = \exp(\bar{h}^{m\top}_{i}{\bar{h}^m_{j}}/\tau_{ea})$.

The comprehensive training objective is formulated as: $\mathcal{L}_{ea} = \mathcal{L}_{GMI} + \mathcal{L}_{ECIA} + \mathcal{L}_{IIR}$.
% , integrating our MMKG representation learning framework with the UMAEA~\cite{chen2023rethinking}. 
Note that our \textbf{\ours}~framework can not only function as a standalone model but also enhance other existing methods, providing stable performance improvements in MMEA, as demonstrated in Table~\ref{tab:mmea}.

% ---------------------------------------------------------
% --------------------  Experiments -----------------------
% ---------------------------------------------------------
\vspace{-1pt}
\section{Experiments Setup}
% \subsection{Experiment Setup}
\vspace{-0.5pt}
\vskip -0.05in
% \subsubsection{\textbf{Iterative Training for MMEA}}
\paragraph{Iterative Training for MMEA.}
We employ a probation technique for iterative training, which acts as a buffering mechanism, temporarily storing a cache of mutual nearest entity pairs across KGs from the testing set~\cite{DBLP:conf/coling/LinZWSW022}. Specifically, at every $K_e$ (where $K_e = 5$) epochs, models identify and add mutual nearest neighbor entity pairs from different KGs to a candidate list $\mathcal{N}^{cd}$. An entity pair in $\mathcal{N}^{cd}$ is then added to the training set if it continues to be mutual nearest neighbors for $K_s$ ($=$ $10$) consecutive iterations. This iterative expansion of the training dataset serves as data augmentation in the EA domain, enabling further evaluation of the model's robustness across various scenarios.

\vspace{-1pt}
\vskip -0.05in
% \subsubsection{\textbf{Implementation Details}}\label{sec:detail}
\paragraph{Implementation Details.}\label{sec:detail}
\textbf{\ul{MKGC:}}
\textbf{\textit{(\rmnum{1})}} Following \citet{zhang2024unleashing}, vision encoders $Enc_{v}$ are configured with VGG~\cite{DBLP:journals/corr/SimonyanZ14a} for DBP15K, and BEiT~\cite{DBLP:conf/iclr/Bao0PW22} for MKG-W and MKG-Y. For entities associated with multiple images, the feature vectors of these images are averaged to obtain a singular representation.
\textbf{\textit{(\rmnum{2})}} The head number $N_h$ in MHCA is set to $2$. For entity representation in DBP15K, graph structure embedding $\bar{h}^g$ is used, while for MKG-W and MKG-Y, mean pooling across modality-specific representations $\bar{h}^{avg}$ is employed. This distinction is made due to DBP15K's denser KG and greater absence of modality information compared to MKG-W and MKG-Y.
\textbf{\textit{(\rmnum{3})}}  We simply selected a set of candidate parameters in AdaMF~\cite{zhang2024unleashing}. Specifically, the number of negative samples $K$ per positive triple is $32$, the hidden dimension $d$ is $256$, the training batch size is $1024$, the margin $\lambda$ is $12$, the temperature $\tau_{kgc}$ is $2.0$, and the learning rate is set to $1e-4$.  No extensive parameter tuning was conducted; theoretically, \ours~could achieve better performance with parameter optimization.
\textbf{\textit{(\rmnum{4})}}
The probability $\rho$ of applying noise in GMNM is set at $0.2$, with a noise ratio $\epsilon$ of $0.7$.
\textbf{\textit{(\rmnum{5})}}
For fairness in comparison, we excluded Ensemble-methods like MoSE~\cite{DBLP:conf/emnlp/ZhaoCWZZZJ22-MOSE} and PLM-based methods like MKGformer~\cite{DBLP:conf/sigir/ChenZLDTXHSC22-MKGformer}  due to significant parameter size differences (our model: 13M; MKGformer: over 200M). 

\input{tab/mmea-tb}
\textbf{\ul{MMEA:}} 
\textbf{\textit{(\rmnum{1})}} Following \citet{DBLP:conf/emnlp/YangZSLLS19}, Bag-of-Words (BoW) is employed for encoding relations ($x^r$) and attributes ($x^a$) into fixed-length vectors ($d_r=d_a=1000$). This process entails sorting relations and attributes by frequency, followed by truncation or padding to standardize vector lengths, thus streamlining representation and prioritizing significant features. For any entity $e_i$, vector positions correspond to the presence or frequency of top-ranked attributes and relations, respectively.
\textbf{\textit{(\rmnum{2})}} Following \cite{DBLP:conf/ksem/ChenLWXWC20,DBLP:conf/coling/LinZWSW022}, vision encoders $Enc_{v}$ are selected as ResNet-152 \cite{he2016deep} for DBP15K, and CLIP \cite{DBLP:conf/icml/RadfordKHRGASAM21} for Multi-OpenEA.
% To focus on the intrinsic alignment capabilities of EA models, we omit textual surface information, following~\cite{chen2023rethinking,wang2024towards}.
\textbf{\textit{(\rmnum{3})}} An alignment editing method is applied to minimize error accumulation \cite{DBLP:conf/ijcai/SunHZQ18}.
\textbf{\textit{(\rmnum{4})}} The head number $N_h$ in MHCA is set to $1$. The hidden layer dimensions $d$ for all networks are unified into $300$.  
The total epochs for baselines are set to $500$ with an option for an additional $500$ epochs of iterative training  \cite{DBLP:conf/coling/LinZWSW022}.
Our training strategies incorporates a cosine warm-up schedule ($15\%$ of steps for LR warm-up), early stopping, and gradient accumulation, using the AdamW optimizer  ($\beta_1=0.9$, $\beta_2=0.999$) with a consistent batch size of $3500$.
\textbf{\textit{(\rmnum{5})}} The total learnable parameters of our model are comparable to those of baseline models. For instance, under the DBP15K$_{{\text{JA-EN}}}$ dataset: EVA has $13.27$M, MCLEA has $13.22$M, and our \ours~has $13.82$M learnable parameters.

\vspace{-1pt}
\vskip -0.05in
% --------------------  Main -----------------------
\section{Experimental Results} 
\vskip -0.05in
% \subsubsection{\textbf{MKGC Results}}
\paragraph{Overall MKGC Results.}
As shown in Tab.~\ref{tab:mkgc}, \ours~achieves SOTA performance across all metrics on three MKGC datasets, especially notable when compared with recent works like MANS~\cite{DBLP:journals/corr/abs-2304-11618-MANS} and MMRNS~\cite{DBLP:conf/mm/Xu0WZC22} which all have refined the Negative Sampling techniques. Our Entity-level Modality Interaction approach for MMKG representation learning not only demonstrates a significant advantage but also benefits from the consistent performance enhancement provided by our Gauss Modality Noise Masking (GMNM) module, maintaining superior performance even in its absence.

\vspace{-1pt}
% \subsubsection{\textbf{MMEA Results}} \label{sec:mmeares}
\paragraph{Overall MMEA Results.}\label{sec:mmeares}
As illustrated in the third segment of Tab.~\ref{tab:mmea}, our \ours~ achieves SOTA performance across all metrics on seven standard MMEA datasets. Notably, in the latter four datasets of the OpenEA series (EN-FR-15K, EN-DE-15K, D-W-15K-V1, D-W-15K-V2) under the \textit{Standard} setting where $R_{img}=1.0$ indicating full image representation for each entity, our GMNM module maintains or even boosts performance. This suggests that strategic noise integration can lead to beneficial results, demonstrating the module's effectiveness even in scenarios where visual data is abundant and complete.
This aligns with findings from related work \cite{chen2023rethinking,chen2023meaformer}, which suggest  that image ambiguities and multi-aspect visual information can sometimes misguide the use of MMKGs. Unlike these studies that typically design models to refuse and combat noise, our \ours~accepts and intentionally integrates noise to better align with the inherently noisy conditions of real-world scenarios.
Iterative training results further confirm the robustness of our approach as detailed in Appendix~\ref{sec:iterT}.

Most importantly, as a versatile MMKG representation learning approach, it is compatible with both MMEA and MKGC tasks, illustrating its robust adaptability in diverse operational contexts.

\input{tab/ablation-tb}

\vspace{-2pt}
% --------------------  Experiments -----------------------
\paragraph{Uncertainly Missing Modality.}
The first two segments from Tab.~\ref{tab:mmea} present entity alignment performance with $R_{img}={0.4, 0.6}$, where 60\%/40\% of entities lack image data. These missing images are substituted with random image features following a normal distribution based on the observed mean and standard deviation across other entities' images (details in \ref{sec:img}). This simulates uncertain modality absence in real-world scenarios. Our method outperforms baselines more significantly when the modality absence is greater (i.e., $R_{img}={0.4}$), with the GMNM module providing notable benefits. This demonstrates that intentionally introducing noise can increase training challenges while enhancing model robustness in realistic settings.

\vspace{-4pt}
% --------------------  Ablation study -----------------------
\paragraph{Ablation studies.}
% In Table~\ref{tab:ablation}, 我们展示了不同 Component对于模型性能的影响。这一过程共分为三个方面：
In Table~\ref{tab:ablation}, we dissect the influence of various components on our model's performance, focusing on three key aspects:
\textbf{\textit{(\rmnum{1})}}  \textbf{Noise Parameters:} The noise application probability $\rho$ and noise ratio $\epsilon$ are pivotal. Optimal values of $\rho = 0.2$ and $\epsilon = 0.7$ were determined empirically, suggesting that the model tolerates up to 20\% of entities missing images and that a modality-mask ratio of $0.7$ acts as a soft mask. For optimal performance, we recommend empirically adjusting these parameters to suit other specific scenario. Generally, conducting a grid search on a smaller dataset subset can quickly identify suitable parameter combinations.
\textbf{\textit{(\rmnum{2})}}  \textbf{Entity-Level Modality Interaction:} Our exploration shows that absence of image information (w/ Only $h^g$) markedly reduces performance, emphasizing MKGC's importance; Weighted summing methods (WS, AT, TS) surpass simple FC-based approaches, indicating the superiority of nuanced modality integration; Using purely Transformer modality weights $\tilde{w}^m$ for weighting does not demonstrate a clear advantage over attention-based or globally learnable weight methods in MKGC. In contrast, our approach, which utilizes $\bar{h}^g$ (for DBP15K) and $\bar{h}^{avg}$ (for MKG-W and MKG-Y), significantly outperforms others, demonstrating its efficacy.
\textbf{\textit{(\rmnum{3})}} \textbf{Modality-Mask vs. Dropout:} In assessing their differential impacts, we observe that even minimal dropout ($0.1$) adversely affects performance, likely because dropout to some extent distorts the original modal feature distribution, thereby hindering model optimization toward the alignment objective. Conversely, our modality-mask's noise is inherent, replicating the feature distribution seen when modality is absent, and consequently enhancing model robustness more effectively.
% ---------------------------------------------------------
% --------------------  Conclusion -----------------------
% ---------------------------------------------------------
\vspace{-4pt}
\section{Conclusion}
\vspace{-4pt}
In this work, we introduce a unified noise-powered multi-modal knowledge graph representation framework that accepts and intentionally integrates noise, thereby aligning with the complexities of real-world scenarios. This initiative also stands out as the first in the MMKG domain to support both MKGC and MMEA tasks simultaneously, highlighting the adaptability of our approach.

\section*{Acknowledgments}
This work is founded by National Natural Science Foundation of China (NSFCU23B2055/NSFCU19B2027/NSFC62306276), Zhejiang Provincial Natural Science Foundation of China (No. LQ23F020017), Yongjiang Talent Introduction Programme (2022A-238-G), and Fundamental Research Funds for the Central Universities (226-2023-00138). This work was supported by AntGroup.

\section*{Limitations}

\paragraph{References \& Definition.}
To aid quick comprehension of tasks within limited space, definitions and boundaries may lack full accuracy and completeness. Detailed explanations and related work are provided in the Appendix~\ref{sec:relat-sup} to elaborate on these concepts.

\paragraph{Benchmarks \& Baselines.}
Due to page constraints, we selected datasets and benchmarks (e.g., Tab.~\ref{tab:mkgc} and Tab.~\ref{tab:mmea}) primarily from recent mainstream works, such as DB15K~\cite{DBLP:conf/esws/LiuLGNOR19}, MKG-W and MKG-Y~\cite{DBLP:conf/mm/Xu0WZC22}. This selection may overlook older datasets like FB15K-237 \cite{DBLP:conf/emnlp/ToutanovaCPPCG15}, WN18 \cite{DBLP:conf/nips/BordesUGWY13}, and WN9-IMG \cite{DBLP:conf/ijcai/XieLLS17}.

For fairness in comparison, we excluded methods based on MoE or Ensemble approaches, such as MoSE~\cite{DBLP:conf/emnlp/ZhaoCWZZZJ22-MOSE}, and did not compare with PLM-based methods like MKGformer~\cite{DBLP:conf/sigir/ChenZLDTXHSC22-MKGformer}  due to significant differences in parameter sizes (our model has only 13M parameters versus MKGformer's over 200M). 

\paragraph{Future Applications.}
Our framework proposes a unified approach for MMKG representation learning, ideally positioned as an MMKG encoder for integrating into LLM training processes, potentially enhancing multi-modal entity embeddings. While our method theoretically supports diverse training objectives, due to the focused scope of this study, we did not validate this aspect experimentally. As the field progresses, we envision further integration of this unified framework into multi-modal knowledge pre-training, potentially supporting various downstream tasks like Multi-modal Knowledge Injection and Retrieval-Augmented Generation (RAG). Such developments could significantly benefit the community, particularly with the rapid advancements in Large Language Models~\cite{DBLP:journals/corr/abs-2402-05391,DBLP:journals/corr/abs-2410-07526}.

\bibliography{custom}

\clearpage
\appendix

\section{Appendix}

\subsection{Supplementary for Related Work}\label{sec:relat-sup}
Typically, a KG is considered multi-modal when it contains knowledge symbols expressed across various modalities, including, but not limited to, text, images, sound, or video. 
Current research primarily concentrates on the visual modality, assuming that other modalities can be processed similarly.

\vspace{-2pt}
\paragraph{Multi-Modal Knowledge Graph Completion.}
Multi-modal Knowledge Graph Completion (MKGC)~\cite{DBLP:journals/corr/abs-2405-16869,DBLP:journals/corr/abs-2404-09468} is crucial for inferring missing triples in existing MMKGs, involving three sub-tasks: Entity Prediction, Relation Prediction, and Triple Classification. Currently, most research in MKGC focuses on Entity Prediction, also widely recognized as Link Prediction, with two main methods emerging: 
\textbf{\ul{Embedding-based Approaches}}
build on conventional Knowledge Graph Embedding (KGE) methods \cite{DBLP:conf/nips/BordesUGWY13,DBLP:conf/iclr/SunDNT19}, adapted to integrate multi-modal data, enhancing entity embeddings.
\textbf{\textit{(\rmnum{1})}} \textbf{Modality Fusion Methods}~\cite{ DBLP:journals/corr/abs-2309-01169-EEMMKG, DBLP:journals/inffus/WangYCSL22, DBLP:journals/corr/abs-2206-13163} integrate multi-modal and structural embeddings to assess triple plausibility. Early efforts, like IKRL~\cite{DBLP:conf/ijcai/XieLLS17-IKRL}, utilize multiple TransE-based scoring functions~\cite{DBLP:conf/nips/BordesUGWY13} for modal interaction. 
RSME~\cite{DBLP:conf/mm/WangWYZCQ21-RSME} employs gates for selective modal information integration. 
OTKGE~\cite{DBLP:conf/nips/CaoXYHCH22-OTKGE} leverages optimal transport for fusion, while CMGNN \cite{DBLP:journals/tkde/FangZHWX23-CMGNN} implements a multi-modal GNN with cross-modal contrastive learning.
HRGAT~\cite{DBLP:journals/tomccap/LiangZZ023-HRGAT} creates a hyper-node relational graph, 
CamE \cite{DBLP:conf/icde/XuZXXLCD23-CamE} focuses on biological KGs with a triple co-attention module, 
VISITA \cite{DBLP:conf/emnlp/LeeCLJW23-VISITA} utilizes a transformer framework for relation and triple-level multi-modal information fusion.
\textbf{\textit{(\rmnum{2})}}~\textbf{Modality Ensemble Methods} train distinct models per modality, merging outputs for predictions. For example, MoSE~\cite{DBLP:conf/emnlp/ZhaoCWZZZJ22-MOSE} utilizes structural, textual, and visual data to train three KGC models and employs, using ensemble strategies for joint predictions. Similarly, IMF~\cite{DBLP:conf/www/LiZXZX23-IMF} proposes an interactive model to achieve modal disentanglement and entanglement to make robust predictions. 
\textbf{\textit{(\rmnum{3})}} \textbf{Modality-aware Negative Sampling Methods}~\cite{DBLP:journals/apin/LuWJHL22-MMKRL,DBLP:journals/corr/abs-2209-07084-VBKGC,DBLP:journals/corr/abs-2304-11618-MANS,DBLP:conf/mm/Xu0WZC22} boost differentiation between correct and erroneous triples by incorporating multi-modal context for superior negative sample selection. 
MMKRL~\cite{DBLP:journals/apin/LuWJHL22-MMKRL} introduces adversarial training to MKGC, adding perturbations to modal embeddings. Following this, VBKGC~\cite{DBLP:journals/corr/abs-2209-07084-VBKGC} and MANS~\cite{DBLP:journals/corr/abs-2304-11618-MANS} develop fine-grained visual negative sampling to better align visual with structural embeddings for more nuanced comparison training. MMRNS~\cite{DBLP:conf/mm/Xu0WZC22} enhances this with relation-based sample selection.
\textbf{\ul{Finetune-based Approaches}}~\cite{DBLP:conf/sigir/ChenZLDTXHSC22-MKGformer,DBLP:journals/corr/abs-2307-03591}
exploit the world understanding capabilities of pre-trained Transformer models like BERT \cite{DBLP:conf/naacl/DevlinCLT19} and VisualBERT \cite{DBLP:journals/corr/abs-1908-03557} for MKGC. 
These approaches reformat MMKG triples as token sequences for PLM processing~\cite{DBLP:journals/corr/abs-2212-05767}, often framing KGC as a classification task.
For example,
MKGformer~\cite{DBLP:conf/sigir/ChenZLDTXHSC22-MKGformer} integrates multi-modal fusion at multiple levels, treating MKGC as a Masked Language Modeling (MLM) task, while 
SGMPT~\cite{DBLP:journals/corr/abs-2307-03591} extends this by incorporating structural data and a dual-strategy fusion module.

\vspace{-2pt}
\paragraph{Multi-Modal Entity Alignment.}
Entity Alignment (EA) is pivotal for KG integration, aiming to identify identical entities across different KGs by leveraging relational, attributive, and literal (surface) features. Multi-Modal Entity Alignment (MMEA) enhances this process by incorporating visual data, thereby improving alignment accuracy accuracy~\cite{DBLP:conf/esws/LiuLGNOR19,DBLP:conf/ksem/ChenLWXWC20,DBLP:conf/mm/NiXJCCH23,DBLP:conf/mm/XuXS23}. 
Introduced in 2020, MMEA~\cite{DBLP:conf/ksem/ChenLWXWC20} merges multiple modalities to align entities in MMKGs by minimizing the distance between their holistic embeddings.
{HMEA} \cite{DBLP:journals/ijon/GuoTZZL21} represents MMKGs on the hyperbolic manifold, offering refined geometric interpretations.
{EVA} \cite{DBLP:conf/aaai/0001CRC21} applies an attention mechanism to modulate the importance of each modality and introduces an unsupervised approach that utilizes visual similarities for alignment, reducing reliance on gold-standard labels.
{MSNEA} \cite{DBLP:conf/kdd/ChenL00WYC22} leverages visual cues to guide relational feature learning. 
{MCLEA} \cite{DBLP:conf/coling/LinZWSW022} employs KL divergence to mitigate the modality distribution gap between uni-modal and joint embeddings.
% {PathFusion}~\cite{DBLP:journals/corr/abs-2310-05364} and ASGEA~\cite{DBLP:journals/corr/abs-2402-11000} combine information from different modalities using the modality similarity or alignment path as an information carrier.
{DFMKE}~\cite{DBLP:journals/inffus/ZhuHM23} employs a late fusion approach with modality-specific low-rank factors that enhance feature integration across various knowledge spaces, complementing early fusion output vectors.
MEAformer~\cite{chen2023meaformer} adjusts mutual modality preferences dynamically for entity-level modality fusion, addressing inconsistencies in entities' surrounding modalities. 
{MoAlign}~\cite{DBLP:journals/corr/abs-2310-06365}, UMAEA~\cite{chen2023rethinking} PCMEA~\cite{DBLP:conf/aaai/WangQBZQ24} and DESAlign~\cite{wang2024towards} follow similar settings.

\vspace{-2pt}
\paragraph{A-MMKG vs. N-MMKG.}
Drawing on the categorization proposed in~\cite{DBLP:journals/corr/abs-2402-05391}, we distinguish between two types of MMKGs: A-MMKG and N-MMKG. In A-MMKGs, images are attached to entities as attributes, while in N-MMKGs, images are treated as standalone entities interconnected with others. A-MMKGs are more prevalent in current research and applications within the semantic web community due to their accessibility and similarity to traditional KGs.  Therefore, this paper will focus exclusively on A-MMKG, unless stated otherwise. For instance, in an MMKG, an attribute triple $(e, a, v)$ in $\mathcal{T_A}$ might associates an image as $v$ to an entity $e$ via an attribute $a$, typically denoted as \textit{hasImage}.

\subsection{Supplementary for Experiments}
\subsubsection{Datasets}
% In MMKG datasets, entities often have associated images, but the proportion ($R_{img}$) of image-containing entities varies, as seen with 67.58$\%$ in DBP15K$_{{\text{JA-EN}}}$. This variability is an inherent aspect of the data collection process in existing MMKG datasets~\cite{DBLP:journals/corr/abs-2402-05391}.
In MMKG datasets like DBP15K${{\text{JA-EN}}}$, where 67.58$\%$ of entities have images, the image association ratio ($R_{img}$) varies due to the data collection process~\cite{chen2023meaformer}.

% \textbf{\ul{MKGC:}}
\paragraph{MKGC:}
\textbf{\textit{(\rmnum{1})}} DB15K~\cite{DBLP:conf/esws/LiuLGNOR19} is constructed from DBPedia~\cite{DBLP:journals/semweb/LehmannIJJKMHMK15}, enriched with images obtained via a search engine.
\textbf{\textit{(\rmnum{2})}}  MKG-W and MKG-Y~\cite{DBLP:conf/mm/Xu0WZC22}  are subsets of Wikidata~\cite{DBLP:journals/cacm/VrandecicK14} and YAGO~\cite{DBLP:conf/www/SuchanekKW07} respectively.
Text descriptions are aligned with the corresponding entities using the additional \textit{sameAs} links provided by OpenEA benchmarks~\cite{DBLP:journals/pvldb/SunZHWCAL20}.
Detailed statistics are available in Tab.~\ref{tab:EAdata}~\&~\ref{table:KGCdata}.
% \textbf{\ul{MMEA:}} 
\input{tab/sta-tab}
\paragraph{MMEA:}
\textbf{\textit{(\rmnum{1})}} Multi-modal DBP15K~\cite{DBLP:conf/aaai/0001CRC21} extends DBP15K~\cite{DBLP:conf/semweb/SunHL17} by adding images from DBpedia and Wikipedia~\cite{denoyer2006wikipedia}, covering three bilingual settings (DBP15K$_{{\text{ZH-EN}}}$, DBP15K$_{{\text{JA-EN}}}$, DBP15K$_{{\text{FR-EN}}}$) and featuring around $400$K triples and $15$K aligned entity pairs per setting.
\textbf{\textit{(\rmnum{2})}} MMEA-UMVM~\cite{chen2023rethinking} includes two bilingual datasets (EN-FR-15K, EN-DE-15K) and two monolingual datasets (D-W-15K-V1, D-W-15K-V2) derived from Multi-OpenEA datasets ($R_{sa}=0.2$)~\cite{DBLP:journals/corr/abs-2302-08774} and all three bilingual datasets from DBP15K~\cite{DBLP:conf/aaai/0001CRC21}. It offers variability in visual information by randomly removing images, resulting in 97 distinct dataset splits with different $R_{img}$. For this study, we focus on representative $R_{img}$ values of $\{0.4$, $0.6$, $maximum\}$ to validate our experiments.
When $R_{img} = maximum$, the dataset corresponds to the original \textit{Standard} dataset (as shown in Tab.~\ref{tab:mmea}). Note that for the Multi-modal DBP15K dataset, the ``$maximum$'' value is not 1.0.

% \subsubsection{Baselines}
% % 需要注意的是对于所有Baseline，我们为了公平对比均选择的end-2-end方法，即没有考虑一些额外的生成流程，比如AdaMF的对抗训练阶段。

\subsubsection{Iterative Training}\label{sec:iterT}
Iterative training results further confirm the robustness of our approach, as shown in Tab.~\ref{tab:mmea-ap}.

\input{tab/mmea-ap-tab}

\subsection{Metric Details}

\subsubsection{MMEA Metrics}
\textbf{\textit{(\rmnum{1})}} \textbf{\ul{MRR}}
% \subsubsection{\textbf{MRR}} 
(Mean Reciprocal Ranking $\uparrow$) is a statistic measure for evaluating many algorithms that produce a list of possible responses to a sample of queries, ordered by probability of correctness. 
In the field of EA, the reciprocal rank of a query entity (i.e., an entity from the source KG) response is the multiplicative inverse of the rank of the first correct alignment entity in the target KG.
MRR is the average of the reciprocal ranks of results for a sample of candidate alignment entities:
\begin{equation}
    \vspace{-2pt}
    \mathbf{MRR}=\frac{1}{|\mathcal{S}_{te}|} \sum_{i=1}^{|\mathcal{S}_{te}|} \frac{1}{\text {rank}_i} \,.
\end{equation}

\textbf{\textit{(\rmnum{2})}} \textbf{\ul{Hits$@$N}}
% \subsubsection{\textbf{Hits$@$N}} 
describes the fraction of true aligned 
 target entities that appear in the first N entities of the sorted rank list:
\begin{equation}
  \vspace{-2pt}
    \operatorname{Hits} @ \text{N}=\frac{1}{|\mathcal{S}_{te}|} \sum_{i=1}^{|\mathcal{S}_{te}|} \mathbb{I}[{\text {rank}_i} \leqslant \text{N}]\, ,
\end{equation}
where  ${{\text{rank}}_{i}}$ refers to the rank position of the first correct mapping for the i-th query entities and $\mathbb{I}=1$ if ${\text {rank}_i} \leqslant N$ and 0 otherwise.
$\mathcal{S}_{te}$ refers to the testing alignment set.

\subsubsection{MKGC Metrics}
MKGC involves predicting the missing entity in a query, either $(h,r,?)$ for tail prediction or $(?,r,t)$ for head prediction. To evaluate the performance, we use rank-based metrics such as mean reciprocal rank (MRR) and Hit@N (N=1, 3, 10), following standard practices in the field.
\textbf{\textit{(\rmnum{1})}} \textbf{\ul{MRR}} is calculated as the average of the reciprocal ranks of the correct entity predictions for both head and tail predictions across all test triples:
\begin{equation}
   \mathbf{MRR}=\frac{1}{|\mathcal{T}_{test}|}\sum_{i=1}^{|\mathcal{T}_{test}|}(\frac{1}{r_{h,i}}+\frac{1}{r_{t,i}})\,.
\end{equation}

\textbf{\textit{(\rmnum{2})}} \textbf{\ul{Hits$@$N}} measures the proportion of correct entity predictions ranked within the top N positions for both head and tail predictions:
\begin{equation}
   \mathbf{Hit@N}=\frac{1}{|\mathcal{T}_{test}|}\sum_{i=1}^{|\mathcal{T}_{test}|}(\mathbb{I}(r_{h,i} \leqslant N)+\mathbb{I}(r_{t,i} \leqslant N))\,,
\end{equation}
where $r_{h,i}$ and $r_{t,i}$ denote the rank positions in head and tail predictions, respectively.

% Besides, filter setting \cite{DBLP:conf/nips/BordesUGWY13} is applied to all the results to avoid the influence of the training triples for fair comparisons.
Additionally, we employ a filter setting~\cite{DBLP:conf/nips/BordesUGWY13} to remove known triples from the ranking process, ensuring fair comparisons and mitigating the impact of known information from the training set on the evaluation metrics.

\end{document}

%% file: tab/mkgc-tb.tex
\begin{table*}[!tbp]
   \vspace{-2pt}
   \vskip -0.15in
\caption{MKGC performance on DB15K~\cite{DBLP:conf/esws/LiuLGNOR19}, MKG-W and MKG-Y~\cite{DBLP:conf/mm/Xu0WZC22} datasets. The best results are highlighted in \textbf{bold}, and the third-best results are \ul{underlined} for each column. 
% MMRNS~\cite{DBLP:conf/mm/Xu0WZC22} uses RotatE by default, while ``$\star$'' represents its best results with different scoring functions. 
% The icon {\small \faToggleOn} signifies the use of Negative Sampling techniques; {\small \faToggleOff} indicates its absence.
}
\centering
% \vspace{-2pt}
   \vskip -0.05in
\resizebox{0.93\textwidth}{!}{
\begin{NiceTabular}{rcccc@{\hspace{20pt}}cccc@{\hspace{20pt}}cccc}
 \CodeBefore
 \rowcolors{2}{gray!10}{white}
 % \columncolor{gray!1}{1}
 \rowcolor{gray!30}{1}
 \rowcolor{gray!30}{2}
 \rowcolor{gray!1}{16}
 \rowcolor{gray!1}{17}
 \Body
 \toprule[0.8pt]
 \multirow{3}*{\makebox[5cm][c]{\textbf{Models}}} & \multicolumn{4}{c@{\hspace{20pt}}}{\textbf{DB15K}~\cite{DBLP:conf/esws/LiuLGNOR19}} & \multicolumn{4}{c@{\hspace{20pt}}}{\textbf{MKG-W}~\cite{DBLP:conf/mm/Xu0WZC22}} & \multicolumn{4}{c}{\textbf{MKG-Y}~\cite{DBLP:conf/mm/Xu0WZC22}} \\
 \cmidrule(l{5pt}r{20pt}){2-5} \cmidrule(r{20pt}){6-9} \cmidrule(r){10-13} 
 & {\footnotesize \textbf{MRR}} & {\footnotesize \footnotesize \textbf{H@1}} & {\footnotesize \textbf{H@3}} & {\footnotesize \textbf{H@10}} & {\footnotesize \textbf{MRR}} & {\footnotesize \footnotesize \textbf{H@1}} & {\footnotesize \textbf{H@3}} & {\footnotesize \textbf{H@10}} & {\footnotesize \textbf{MRR}} & {\footnotesize \footnotesize \textbf{H@1}} & {\footnotesize \textbf{H@3}} & {\footnotesize \textbf{H@10}}\\
 \midrule[0.8pt]
 IKRL {\footnotesize {(IJCAI '17)~\cite{DBLP:conf/ijcai/XieLLS17-IKRL}}}  & .268 & .141 & .349 & .491 & .324 & .261 & .348 & .441 & .332 & .304 & .343 & .383 \\
 TBKGC {\footnotesize (NAACL '18)~{\cite{DBLP:conf/starsem/SergiehBGR18-TBKGC}}}& .284 & .156 & .370 & .499 & .315 & .253 & .340 & .432 & .340 & .305 & .353 & .401 \\
 TransAE {\footnotesize (IJCNN '19)~{\cite{DBLP:conf/ijcnn/WangLLZ19-TransAE}}} & .281 & .213 & .312 & .412 & .300 & .212 & .349 & .447 & .281 & .253 & .291 & .330 \\
 % MMKRL {\footnotesize (Appl. Intell. '22)~{\cite{DBLP:journals/apin/LuWJHL22-MMKRL}}}& .268 & .139 & .351 & .494 & .301 & .222 & .341 & .447 & .368 & .317 & .398 & .453 \\
 RSME {\footnotesize (ACM MM '21)~{\cite{DBLP:conf/mm/WangWYZCQ21-RSME}}} & .298 & \ul{.242} & .321 & .403 & .292 & .234 & .320 & .404 & .344 & .318 & .361 & .391 \\
 VBKGC {\footnotesize (KDD '22)~{\cite{DBLP:journals/corr/abs-2209-07084-VBKGC}}}& .306 & .198 & .372 & .494 & .306 & .249 & .330 & .409 & .370 & \ul{.338} & .388 & .423 \\
 OTKGE {\footnotesize (NeurIPS '22)~{\cite{DBLP:conf/nips/CaoXYHCH22-OTKGE}}}& .239 & .185 & .259 & .342 & .344 & \ul{.289} & .363 & .449 & .355 & .320 & .372 & .414 \\
% ------
 IMF {\footnotesize (WWW '23)~{\cite{DBLP:conf/www/LiZXZX23-IMF}}} & .323 & .242 & .360 & .482 & \ul{.345} & .288 & .366 & .454 & .358 & .330 & .371 & .406 \\
 QEB {\footnotesize (ACM MM '23)~{\cite{wang2023tiva}}}& .282 & .148 & .367 & .516 & .324 & .255 & .351 & .453 & .344 & .295 & .370 & .423 \\
 VISTA {\footnotesize (EMNLP '23)~{\cite{DBLP:conf/emnlp/LeeCLJW23-VISITA}}}& .304 & .225 & .336 & .459 & .329 & .261 & .354 & .456 & .305 & .249 & .324 & .415 \\
% ------
 MANS {\footnotesize (IJCNN '23)~{\cite{DBLP:journals/corr/abs-2304-11618-MANS}}} & .288 & .169 & .366 & .493 & .309 & .249 & .336 & .418 & .290 & .253 & .314 & .345 \\
 MMRNS {\footnotesize (ACM MM '22)~{\cite{DBLP:conf/mm/Xu0WZC22}}} & .297 & .179 & .367 & .510 & .341 & .274 & .375 & .468 & .359 & .306 & .391 & \ul{.455} \\
%  MMRNS$^\star$~{\footnotesize {\cite{DBLP:conf/mm/Xu0WZC22}}} & \ul{.327} & .230 & .379 & .510 & \ul{.350} & .286 & .375 & \ul{.475} & .359 & .306 & .391 & .455 \\
 AdaMF {\footnotesize (COLING '24)~{\cite{zhang2024unleashing}}} & \ul{.325} & .213 & \ul{.397} & \ul{.517} & .343 & .272 & \ul{.379} & \ul{.472} & \ul{.381} & .335 & \ul{.404} & \ul{.455} \\
\midrule
 \textbf{\ours} {\footnotesize (Ours) \quad} & \textbf{.363} & \textbf{.274} & \textbf{.411}& \textbf{.530}& \textbf{.373}	& \textbf{.302}	& \textbf{.405} &	\textbf{.503} & \textbf{.395}& \textbf{.354} & \textbf{.411}& \textbf{.471}\\
 \quad {\small - w/o  GMNM} & {.357} & {.269} & {.406}& {.523}& {.365} & {.296}	& {.398} &{.490} & {.387}& {.345} & {.407} & {.457}\\
\bottomrule[0.8pt]
\end{NiceTabular}}

   \vspace{-2pt}
   \vskip -0.15in
\label{tab:mkgc}
\end{table*}

%% file: tab/mmea-tb.tex
\begin{table}[!t]
\centering
\tabcolsep=0.3cm
\renewcommand\arraystretch{1.0}
   % \vspace{-5pt}
   % \vskip -0.05in
\caption{{Non-iterative} MMEA results across three degrees of visual modality missing.
% : $R_{img}$ $=$ $\{0.4$, $0.6$, $maximum\}$.  
Results are \underline{underlined} when the baseline, equipped with the Gauss Modality Noise Masking (GMNM) module, surpasses its own original performance, and highlighted in \textbf{bold} when achieving SOTA performance.
} 
\vspace{-2pt}
   \vskip -0.05in
\resizebox{1.0\linewidth}{!}{
\setlength{\tabcolsep}{4.5pt} % 将列间距设置为2pt
\begin{NiceTabular}{@{}l|l|ccc|ccc|ccc}
\CodeBefore
 \rowcolors{2}{gray!10}{white}
 \columncolor{gray!1}{1}
 \rowcolor{gray!30}{1}
 \rowcolor{gray!30}{2}
 \Body
 \toprule[0.8pt]
& \multirow{2}*{\makebox[1.6cm][c]{\textbf{Models}}} & \multicolumn{3}{c|}{\bm{$R_{img}$ $=$ $0.4$}} & \multicolumn{3}{c|}{\bm{$R_{img}$ $=$ $0.6$}} & \multicolumn{3}{c}{\textbf{Standard}} \\
& &  {\footnotesize \textbf{H@1}} & {\footnotesize \textbf{H@10}} & {\footnotesize \textbf{MRR}} & {\footnotesize \textbf{H@1}} & {\footnotesize \textbf{H@10}} & {\footnotesize \textbf{MRR}} & {\footnotesize \textbf{H@1}} & {\footnotesize \textbf{H@10}} & {\footnotesize \textbf{MRR}} \\
 \midrule[0.8pt]
% ------------------------------------ DBP ZH-EN -------------------------------------
\parbox[t]{2mm}{\multirow{7}{*}{\rotatebox[origin=c]{90}{\small DBP15K$_{{\text{ZH-EN}}}$}}} 
& EVA  &
{.623} &{.876} &{.715} & .625 & .877 & .717 & .683 & .906 & .762 \\
 & \quad {\small w/  GMNM} & \ul{.629} & \ul{.883} & \ul{.724} & {.625} & \ul{.881} & {.717} & {.680} & \ul{.907} & {.760} \\ 
& MCLEA   &
 {.627} & {.880} & {.715} &{.670} &{.899} &{.751} & .732 & .926 & .801 \\
 & \quad {\small w/  GMNM} & \ul{.652} & \ul{.895} & \ul{.740} & \ul{.699} & \ul{.912} & \ul{.775} & \ul{.754} & \ul{.933} & \ul{.819} \\ 
& {\small MEAformer}     
&  {.678} & {.924} & {.766} & {.720} & {.938} & {.798} & .776 & .953 & .840  \\
& \quad {\small w/  GMNM} & \ul{.680} & \ul{.925} & \ul{.767} & {.719} & \ul{.939} & {.798} & \ul{.777} & \ul{.955} & \ul{.841} \\ 
% & {UMAEA} {\footnotesize {\cite{chen2023rethinking}}}  
% &  {.715} & {.937} & {.796} & {.749} & {.948} & {.822} & .797 & .960 & .856  \\
& \ours~{\footnotesize (Ours)} & \textbf{.735} & \textbf{.945} & \textbf{.812} & \textbf{.757} & \textbf{.953} & \textbf{.830} & \textbf{.798} & \textbf{.963} & \textbf{.858} \\ 
\midrule
% ------------------------------------ DBP JA-EN -------------------------------------
\parbox[t]{2mm}{\multirow{7}{*}{\rotatebox[origin=c]{90}{\small DBP15K$_{{\text{JA-EN}}}$}}}
& EVA   &{.546} &{.829} &{.644} & .552 & .829 & .647 & .587 & .851 & .678 \\
& \quad {\small w/  GMNM} & \ul{.618} & \ul{.876} & \ul{.709} & \ul{.625} & \ul{.874} & \ul{.714} & \ul{.664} & \ul{.902} & \ul{.748} \\ 
& MCLEA   &
  {.568} & {.848} & {.665} &{.639} &{.882} &{.723} & .678 & .897 & .755 \\
 & \quad {\small w/  GMNM} & \ul{.659} & \ul{.901} & \ul{.745} & \ul{.723} & \ul{.924} & \ul{.795} & \ul{.752} & \ul{.935} & \ul{.818} \\ 
& {\small MEAformer}    
&  {.677} & {.933} & {.768} & {.736} & {.953} & {.815}  & .767 & .959 & .837 \\
& \quad {\small w/  GMNM} & \ul{.678} & \ul{.937} & \ul{.770} & \ul{.738} & {.953} & \ul{.816} & {.767} & {.958} & {.837} \\ 
% & {UMAEA} {\footnotesize {\cite{chen2023rethinking}}}  
% &  {.726} & {.946} & {.806} & {.769} & {.959} & {.839}  & .794 & .962 & .856 \\
&  \ours~{\footnotesize (Ours)} & \textbf{.735} & \textbf{.952} & \textbf{.814} & \textbf{.771} & \textbf{.961} & \textbf{.841} & \textbf{.795} & \textbf{.963} & \textbf{.857} \\ 
\midrule
% ------------------------------------ DBP FR-EN -------------------------------------
\parbox[t]{2mm}{\multirow{7}{*}{\rotatebox[origin=c]{90}{\small DBP15K$_{{\text{FR-EN}}}$}}} 
& EVA   &{.622} &{.895} &{.719} & .634 & .899 & .728 & .686 & .926 & .771 \\
& \quad {\small w/  GMNM} & \ul{.628} & \ul{.897} & \ul{.725} & {.634} & \ul{.900} & {.728} & {.686} & \ul{.929} & \ul{.772} \\ 
& MCLEA   &
  {.622} & {.892} & {.722} &{.694} &{.915} &{.774} & .734 & .926 & .805\\
 & \quad {\small w/  GMNM} & \ul{.663} & \ul{.916} & \ul{.756} & \ul{.726} & \ul{.934} & \ul{.802} & \ul{.759} & \ul{.942} & \ul{.827} \\ 
& {\small MEAformer}    
&  {.676} & {.944} & {.774} & {.734} & {.958} & {.816} & .776 & .967 &  .846 \\
& \quad {\small w/  GMNM} & \ul{.678} & \ul{.946} & \ul{.776} & \ul{.735} & \ul{.965} & \ul{.819} & \ul{.779} & \ul{.969} & \ul{.849} \\
% & {UMAEA} {\footnotesize {\cite{chen2023rethinking}}}  
% &  {.746} & {.958} & {.826} & {.785} & {.965} & {.853} & .811 & .972 &  .872 \\
& \ours~{\footnotesize (Ours)} & \textbf{.757} & \textbf{.963} & \textbf{.835} & \textbf{.790} & \textbf{.970} & \textbf{.858} & \textbf{.814} & \textbf{.974} & \textbf{.875} \\ 
% ------------------------------------ Open EA EN-FR-------------------------------------
\midrule
\parbox[t]{2mm}{\multirow{7}{*}{\rotatebox[origin=c]{90}{\small OpenEA$_{{\text{EN-FR}}}$}}} 
& EVA   
& {.532} & {.830} &{.635} & .553 & .835 & .652 & .784 & .931 & .836 \\
& \quad {\small w/  GMNM} & \ul{.537} & {.829} & \ul{.638} & \ul{.554} & {.833} & {.652} & \ul{.787} & \ul{.935} & \ul{.839} \\ 
& MCLEA   
& {.535} &{.842} & {.641} &{.607} &{.858} &{.696} & .821 & .945 & .866 \\
& \quad {\small w/  GMNM} & \ul{.554} & \ul{.848} & \ul{.658} & \ul{.624} & \ul{.873} & \ul{.714} & \ul{.830} & \ul{.950} & \ul{.874} \\ 
& {\small MEAformer}    
&  {.582} & {.891} & {.690} & {.645} & {.904} & {.737}  & .846 & .862 & .889 \\
& \quad {\small w/  GMNM} & \ul{.588} & \ul{.895} & \ul{.696} & \ul{.647} & \ul{.905} & \ul{.738} & \ul{.847} & \ul{.963} & \ul{.890} \\ 
% & {UMAEA} {\footnotesize {\cite{chen2023rethinking}}}  
% &  {.614} & {.897} & {.716} & {.664} & {.914} & {.754}  & .846 & .864 & .890 \\
& \ours~{\footnotesize (Ours)} & \textbf{.621} & \textbf{.905} & \textbf{.721} & \textbf{.667} & \textbf{.922} & \textbf{.757} & \textbf{.848} & \textbf{.964} & \textbf{.891} \\ 
% ------------------------------------ Open EA EN-DE-------------------------------------
\midrule
\parbox[t]{2mm}{\multirow{7}{*}{\rotatebox[origin=c]{90}{\small OpenEA$_{{\text{EN-DE}}}$}}} 
& EVA   & {.718} &{.918} &{.789} & .734 &{.921} & .800 & .922 & .982 & .945 \\
& \quad {\small w/  GMNM} & \ul{.728} & \ul{.919} & \ul{.794} & \ul{.740} & {.921} & \ul{.803} & \ul{.923} & \ul{.983} & \ul{.946} \\ 
& MCLEA   
& {.702} & {.910} & {.774} &{.748} & {.912} &{.805} & .940 & .988 & .957 \\
& \quad {\small w/  GMNM} & \ul{.711} & \ul{.912} & \ul{.782} & \ul{.762} & \ul{.928} & \ul{.821} & \ul{.942} & \ul{.990} & \ul{.960} \\ 
& {\small MEAformer}    
&  {.749} & {.938} & {.816} & {.789} & {.951} & {.847} & .955 & .994 & .971 \\
& \quad {\small w/  GMNM} & \ul{.753} & \ul{.939} & \ul{.817} & \ul{.791} & \ul{.952} & \ul{.848} & \ul{.957} & \ul{.995} & {.971} \\ 
% & {UMAEA} {\footnotesize {\cite{chen2023rethinking}}}  
% &  {.771} & {.945} & {.833} & {.807} & {.956} & {.860} & .955 & .993 & .970 \\
& \ours~{\footnotesize (Ours)} & \textbf{.776} & \textbf{.948} & \textbf{.837} & \textbf{.810} & \textbf{.958} & \textbf{.862} & \textbf{.958} & {.995} & \textbf{.972} \\ 
% ------------------------------------ Open EA D-W-V1-------------------------------------
\midrule
\parbox[t]{2mm}{\multirow{7}{*}{\rotatebox[origin=c]{90}{\small OpenEA$_{{\text{D-W-V1}}}$}}} 
& EVA   &{.567} & {.796} & {.651} & .592 & .810 & .671 & .859 & .945 & .890 \\
& \quad {\small w/  GMNM} & \ul{.597} & \ul{.826} & \ul{.678} & \ul{.611} & \ul{.826} & \ul{.688} & \ul{.870} & \ul{.953} & \ul{.900} \\
& MCLEA   & {.586} &{.821} &{.672} &{.663} &{.854} &{.732} & .882 & .955 & .909 \\
& \quad {\small w/  GMNM} & \ul{.604} & \ul{.841} & \ul{.689} & \ul{.678} & \ul{.869} & \ul{.748} & \ul{.889} & \ul{.960} & \ul{.915} \\
& {\small MEAformer}    
& {.640} & {.877} & {.725} & {.706} & {.898} & {.776} & .902 & .969 & .927 \\
& \quad {\small w/  GMNM} & \ul{.656} & \ul{.884} & \ul{.738} & \ul{.718} & \ul{.905} & \ul{.786} & \ul{.904} & \ul{.971} & \ul{.929} \\
% & {UMAEA} {\footnotesize {\cite{chen2023rethinking}}}  
% &  {.668} & {.890} & {.750} & {.722} & {.908} & {.791} & .905 & .972 & .930 \\
& \ours~{\footnotesize (Ours)} & \textbf{.678} & \textbf{.897} & \textbf{.758} & \textbf{.728} & \textbf{.915} & \textbf{.796} & \textbf{.905} & {.971} & \textbf{.930} \\
% ------------------------------------ Open EA D-W-V2-------------------------------------
\midrule
\parbox[t]{2mm}{\multirow{7}{*}{\rotatebox[origin=c]{90}{\small OpenEA$_{{\text{D-W-V2}}}$}}} 
& EVA   & {.774} &{.949} &{.838} & .789 &{.953} & .848 & .889 & .981 & .922 \\
& \quad {\small w/  GMNM} & \ul{.787} & \ul{.956} & \ul{.848} & \ul{.799} & \ul{.958} & \ul{.856} & \ul{.892} & \ul{.983} & \ul{.924} \\
& MCLEA   &
  {.751} & {.941} & {.822} &{.801} & {.950} &{.856} & .929 & .984 & .950 \\
& \quad {\small w/  GMNM} & \ul{.766} & \ul{.956} & \ul{.836} & \ul{.811} & \ul{.965} & \ul{.868} &\ul{.938} & \ul{.990} & \ul{.957} \\
& {\small MEAformer}   
&  {.807} & {.976} & {.869} & {.834} & {.980} & {.886}  & .939 & .994 &  .960 \\   
& \quad {\small w/  GMNM} & \ul{.833} & \ul{.980} & \ul{.886} & \ul{.857} & \ul{.983} & \ul{.903} & \ul{.942} & \ul{.995} & \ul{.962} \\
% & {UMAEA} {\footnotesize {\cite{chen2023rethinking}}} 
% &  {.846} & {.984} & {.897} & {.858} & {.986} & {.905} & .946 & .996 & .965 \\   
& \ours~{\footnotesize (Ours)} & \textbf{.852} & \textbf{.986} & \textbf{.901} & \textbf{.870} & \textbf{.988} & \textbf{.913} & \textbf{.946} & \textbf{.996} & \textbf{.965} \\
\bottomrule[0.8pt]
\end{NiceTabular}
}
   \vspace{-5pt}
   \vskip -0.15in
\label{tab:mmea}
\end{table}

%% file: tab/ablation-tb.tex
\begin{table}[!t]
   % \vspace{-2pt}
   % \vskip -0.05in
\caption{Component Analysis for \ours~on MKGC datasets.
The icon {\small \faToggleOn}  indicates the activation of the Gauss Modality Noise Masking (GMNM) module; {\small \faToggleOff} denotes its deactivation. By default, GMNM's noise application probability $\rho$ is set to $0.2$, with a noise ratio $\epsilon$ of $0.7$.
Our Transformer-based structure serves as the default fusion method for \ours. Alternatives include: ``FC'' (concatenating features from various modalities followed by a fully connected layer); ``WS'' (summing features weighted by a global learnable weight per modality); ``AT'' (leveraging an Attention network for entity-level weighting); ``TS'' (using a Transformer for weighting to obtain confidence scores $\tilde{w}^m$ for weighted summing); ``w/ Only $h^g$'' (using Graph Structure embedding for uni-modal KGC). ``Dropout'' is an experimental adjustment where  Equation \eqref{eq:1} is replaced with the Dropout function to randomly zero modal input features, based on a defined probability.
}
\vspace{-1pt}
   \vskip -0.05in
\centering
\resizebox{1.0\linewidth}{!}{
\begin{NiceTabular}{@{}lccccccccc@{}}
 \CodeBefore
 \rowcolors{2}{gray!10}{white}
 % \columncolor{gray!1}{1}
 \rowcolor{gray!30}{1}
 \rowcolor{gray!30}{2}
 \Body
 \toprule[0.8pt]
 \multirow{3}*{\makebox[2.5cm][c]{\textbf{Variants}}} & \multicolumn{3}{c}{\textbf{DB15K}} & \multicolumn{3}{c}{\textbf{MKG-W}} & \multicolumn{3}{c}{\textbf{MKG-Y}} \\
 \cmidrule(r){2-4} \cmidrule(r){5-7} \cmidrule(r){8-10} 
 & {\footnotesize \textbf{MRR}} & {\footnotesize \footnotesize \textbf{H@1}} & {\footnotesize \textbf{H@10}} & {\footnotesize \textbf{MRR}} & {\footnotesize \footnotesize \textbf{H@1}} & {\footnotesize \textbf{H@10}} & {\footnotesize \textbf{MRR}} & {\footnotesize \footnotesize \textbf{H@1}} & {\footnotesize \textbf{H@10}}\\
 \midrule[0.8pt]
 {\small \faToggleOn} \textbf{\ours} {\small (Full)}  & \textbf{.363} & \textbf{.274} & \textbf{.530}& \textbf{.373}	& \textbf{.302}	&	\textbf{.503} & \textbf{.395}& \textbf{.354} & \textbf{.471}\\
{\small \faToggleOn} $\rho=0.3$, $\epsilon=0.6$ & .361 & .272 & .528 & .373 & .302 & .502 & .393 & .353 & .468\\
{\small \faToggleOn} $\rho=0.1$, $\epsilon=0.8$ & .360 & .272 & .525 & .371 & .299 & .496 & .391 & .348 & .463\\
{\small \faToggleOn} $\rho=0.4$, $\epsilon=0.4$ & .358 & .268 & .526 & .365 & .296 & .492 & .388 & .346 & .458\\
{\small \faToggleOn} $\rho=0.5$, $\epsilon=0.2$ & .360 & .270 & .528 & .368 & .299 & .493 & .389 & .348 & .457\\
{\small \faToggleOn} $\rho=0.7$, $\epsilon=0.2$ & .359 & .270 & .526 & .367 & .299 & .490 & .387 & .345 & .456\\
 \midrule
{\small \faToggleOff} \textbf{\ours} & {.357} & {.269} & {.523}& {.365} & {.296} &{.490} & {.387}& {.345} & {.457}\\
\midrule
{\small \faToggleOff} - FC Fusion & {.327} & {.210} & {.522}& {.350} & {.287} &{.467} & {.378}& {.340} & {.442}\\
{\small \faToggleOff} - WS Fusion & {.334} & {.218} & {.529}& {.361} & {.298} &{.480} & {.384}& {.345} & {.449}\\
{\small \faToggleOff} - AT Fusion & {.336} & {.225} & {.528}& {.361} & {.296} &{.481} & {.379}& {.343} & {.445}\\
{\small \faToggleOff} - TS Fusion & {.335} & {.221} & {.529}& {.358} & {.292} &{.472} & {.378}& {.344} & {.437}\\
{\small \faToggleOff} - w/ Only $h^g$ & {.293} & {.179} & {.497}& {.337} & {.268} &{.467} & {.350}& {.291} & {.453}\\
\midrule
{\small \faToggleOff} - Dropout (0.1) & {.349} & {.252} & {.527}& {.361} & {.297} &{.479} & {.382}& {.344} & {.446}\\
{\small \faToggleOff} - Dropout (0.2) & {.346} & {.249} & {.526}& {.359} & {.294} &{.478} & {.381}& {.343} & {.446}\\
{\small \faToggleOff} - Dropout (0.3) & {.343} & {.242} & {.524}& {.356} & {.290} &{.477} & {.381}& {.343} & {.445}\\
{\small \faToggleOff} - Dropout (0.4) & {.341} & {.238} & {.521}& {.356} & {.295} &{.467} & {.379}& {.341} & {.442}\\
\bottomrule[0.8pt]
\end{NiceTabular}}
   \vspace{-5pt}
   \vskip -0.15in
\label{tab:ablation}
\end{table}

%% file: tab/sta-tab.tex
\begin{table}[!tbp]
\centering
\caption{Statistics for the MKGC datasets, where the symbol definitions in the table header align with Definition \ref{def:mmkg}.}
\label{table:KGCdata}
\vspace{-2pt}
\renewcommand\arraystretch{1.0}
\setlength{\tabcolsep}{4pt} 
\resizebox{0.84\linewidth}{!}{
\begin{NiceTabular}{cccccc}
\CodeBefore
\rowcolors{2}{gray!10}{white}
\rowcolor{gray!30}{1}
 \Body
\toprule[0.8pt]
Dataset   & $|\mathcal{E}|$    & $|\mathcal{R}|$   & $|\mathcal{T_R}$ {\footnotesize (Train)}$|$ & $|\mathcal{T_R}$ {\footnotesize (Valid)}$|$ & $|\mathcal{T_R}$ {\footnotesize (Test)}$|$  \\ 
\midrule[0.8pt]
DB15K & 12842 & 279 & 79222  & 9902   & 9904  \\
MKG-W    & 15000 & 169  & 34196   & 4276    & 4274   \\
MKG-Y    & 15000 & 28  & 21310   &  2665   & 2663   \\
\bottomrule[0.8pt]
\end{NiceTabular}}
   \vspace{-5pt}
   \vskip -0.15in
\end{table}

\begin{table}[!tbp]
    \centering
    % \vspace{-0.1cm}
    % \footnotesize
    \caption{Statistics for the MMEA datasets. Each dataset contains 15,000 pre-aligned entity pairs ($|\mathcal{S}|=15000$). 
    Note that not every entity is paired with associated images or equivalent counterparts in the other KG. 
    % For dataset \{ EN-FR-15K, EN-DE-15K, D-W-15K-V1, and D-W-15K-V2 \} in Multi-OpenEA, we exclude the ``15K'' suffix for consistency in this paper.  
    Additional abbreviations include: DB (DBpedia), WD (Wikidata), ZH (Chinese), JA (Japanese), FR (French), EN (English), DE (German).}
    \label{tab:EAdata}
    \vspace{-2pt}
    \renewcommand\arraystretch{1.0}
    \setlength{\tabcolsep}{4pt} 
    \resizebox{1.\linewidth}{!}{
    \begin{NiceTabular}{@{}lcccccccc@{}}
     \CodeBefore
     % \rowcolors{2}{gray!10}{white}
     % \columncolor{gray!1}{1}
        \rowcolor{gray!30}{1}
        \rowcolor{gray!10}{2}
        \rowcolor{gray!10}{3}
        \rowcolor{gray!1}{4}
        \rowcolor{gray!1}{5}
        \rowcolor{gray!10}{6}
        \rowcolor{gray!10}{7}
        \rowcolor{gray!1}{8}
        \rowcolor{gray!1}{9}
        \rowcolor{gray!10}{10}
        \rowcolor{gray!10}{11}
        \rowcolor{gray!1}{12}
        \rowcolor{gray!1}{13}
        \rowcolor{gray!10}{14}
        \rowcolor{gray!10}{15}
     \Body
        \toprule[0.8pt]
        \makebox[2cm][c]{Dataset} & $\mathcal{G}$ &  $|\mathcal{E}|$ &  $|\mathcal{R}|$ &  $|\mathcal{A}|$ &  $|\mathcal{T_R}|$ &  $|\mathcal{T_A}|$ &  $|\mathcal{V}_{MM}|$ \\
        \midrule[0.8pt]
        \multirow{2}*{DBP15K$_{{\text{ZH-EN}}}$} & ZH  & 19,388 & 1,701 & 8,111 & 70,414 & 248,035 & 15,912 \\
        & EN  & 19,572 & 1,323 & 7,173 & 95,142 & 343,218 & 14,125 \\
        \midrule
        \multirow{2}*{DBP15K$_{{\text{JA-EN}}}$} & JA  & 19,814 & 1,299 & 5,882 & 77,214 & 248,991 & 12,739  \\
        & EN  & 19,780 & 1,153 & 6,066 & 93,484 & 320,616 & 13,741 \\
        \midrule
        \multirow{2}*{DBP15K$_{{\text{FR-EN}}}$} & FR  & 19,661 & 903 & 4,547 & 105,998 & 273,825 & 14,174  \\
        & EN  & 19,993 & 1,208 & 6,422 & 115,722 & 351,094 & 13,858 \\
        \midrule
        \multirow{2}*{OpenEA$_{{\text{EN-FR}}}$} & EN  & 15,000 & 267 & 308 & 47,334 & 73,121 & 15,000  \\
        & FR  & 15,000 & 210 & 404 & 40,864 & 67,167 & 15,000 \\
        \midrule
        \multirow{2}*{OpenEA$_{{\text{EN-DE}}}$} & EN  & 15,000 & 215 & 286 & 47,676 & 83,755 & 15,000  \\
        & DE & 15,000 & 131 & 194 & 50,419 & 156,150 & 15,000 \\
                \midrule
        \multirow{2}*{OpenEA$_{{\text{D-W-V1}}}$} & DB & 15,000 & 248 & 342 & 38,265 & 68,258 & 15,000  \\
        & WD & 15,000 & 169 & 649 & 42,746 & 138,246 & 15,000 \\
        \midrule
        \multirow{2}*{OpenEA$_{{\text{D-W-V2}}}$} & DB & 15,000 & 167 & 175 & 73,983 & 66,813 & 15,000  \\
        & WD & 15,000 & 121 & 457 & 83,365 & 175,686 & 15,000 \\
        \bottomrule[0.8pt]
    \end{NiceTabular}
    }
    % \vspace{-0.6cm}
    \vspace{-5pt}
   \vskip -0.15in
\end{table}

%% file: tab/mmea-ap-tab.tex
\begin{table}[!t]
\centering
\tabcolsep=0.3cm
\renewcommand\arraystretch{1.0}
\caption{{Iterative} MMEA results.
% across three degrees of visual modality missing: $R_{img}$ $=$ $\{0.4$, $0.6$, $maximum\}$. 
} 
\vspace{-2pt}
\resizebox{1.0\linewidth}{!}{
\setlength{\tabcolsep}{4.5pt} % 将列间距设置为2pt
\begin{NiceTabular}{@{}l|l|ccc|ccc|ccc}
\CodeBefore
 \rowcolors{2}{gray!10}{white}
 \columncolor{gray!1}{1}
 \rowcolor{gray!30}{1}
 \rowcolor{gray!30}{2}
 \Body
 \toprule[0.8pt]
& \multirow{2}*{\makebox[1.6cm][c]{\textbf{Models}}} & \multicolumn{3}{c|}{\bm{$R_{img}$ $=$ $0.4$}} & \multicolumn{3}{c|}{\bm{$R_{img}$ $=$ $0.6$}} & \multicolumn{3}{c}{\textbf{Standard}} \\
& &  {\footnotesize \textbf{H@1}} & {\footnotesize \textbf{H@10}} & {\footnotesize \textbf{MRR}} & {\footnotesize \textbf{H@1}} & {\footnotesize \textbf{H@10}} & {\footnotesize \textbf{MRR}} & {\footnotesize \textbf{H@1}} & {\footnotesize \textbf{H@10}} & {\footnotesize \textbf{MRR}} \\
 \midrule[0.8pt]
% ------------------------------------ DBP ZH-EN -------------------------------------
\parbox[t]{2mm}{\multirow{7}{*}{\rotatebox[origin=c]{90}{\small DBP15K$_{{\text{ZH-EN}}}$}}} 
& EVA  &
{.696} &{.902} &{.773} & .699 & .903 & .775 & .749 & .914 & .810 \\
 & \quad {\small w/  GMNM} & \ul{.708} & \ul{.906} & \ul{.780} & \ul{.705} & \ul{.911} & \ul{.778} & \ul{.752} & \ul{.919} & \ul{.813} \\ 
& MCLEA   &
 {.719} & {.921} & {.796} &{.764} &{.941} &{.831} & .818 & .956 & .871 \\
 & \quad {\small w/  GMNM} & \ul{.741} & \ul{.945} & \ul{.818} & \ul{.782} & \ul{.954} & \ul{.846} & \ul{.830} & \ul{.968} & \ul{.882} \\ 
& {\small MEAformer}    
&  {.754} & {.953} & {.829} & {.788} & {.958} & {.853} & .843 & .966 & .890  \\
& \quad {\small w/  GMNM} & \ul{.763} & \ul{.947} & \ul{.832} & \ul{.799} & \ul{.959} & \ul{.860} & \ul{.845} & \ul{.970} & \ul{.891} \\ 
& \ours~{\footnotesize (Ours)} & \textbf{.798} & \textbf{.957} & \textbf{.859} & \textbf{.821} & \textbf{.963} & \textbf{.876} & \textbf{.857} & \textbf{.972} & \textbf{.900} \\ 
\midrule
% ------------------------------------ DBP JA-EN -------------------------------------
\parbox[t]{2mm}{\multirow{7}{*}{\rotatebox[origin=c]{90}{\small DBP15K$_{{\text{JA-EN}}}$}}}
& EVA   &{.646} &{.888} &{.733} & .657 & .892 & .743 & .695 & .904 & .770 \\
& \quad {\small w/  GMNM} & \ul{.696} & \ul{.910} & \ul{.773} & \ul{.700} & \ul{.912} & \ul{.776} & \ul{.745} & \ul{.916} & \ul{.807} \\ 
& MCLEA   &
  {.690} & {.922} & {.778} &{.756} &{.948} &{.828} & .788 & .955 & .851 \\
 & \quad {\small w/  GMNM} & \ul{.739} & \ul{.937} & \ul{.815} & \ul{.796} & \ul{.959} & \ul{.858} & \ul{.820} & \ul{.969} & \ul{.877} \\ 
& {\small MEAformer}   
&  {.759} & {.957} & {.833} & {.808} & {.969} & {.868}  & .831 & .972 & .882 \\
& \quad {\small w/  GMNM} & \ul{.769} & {.953} & \ul{.838} & \ul{.817} & {.967} & \ul{.872} & \ul{.842} & \ul{.974} & \ul{.890} \\ 
&  \ours~{\footnotesize (Ours)} & \textbf{.808} & \textbf{.959} & \textbf{.864} & \textbf{.839} & \textbf{.975} & \textbf{.890} & \textbf{.861} & \textbf{.976} & \textbf{.904} \\ 
\midrule
% ------------------------------------ DBP FR-EN -------------------------------------
\parbox[t]{2mm}{\multirow{7}{*}{\rotatebox[origin=c]{90}{\small DBP15K$_{{\text{FR-EN}}}$}}} 
& EVA   &{.710} &{.931} &{.792} & .716 & .935 & .797 & .769 & .946 & .834 \\
& \quad {\small w/  GMNM} & \ul{.714} & {.929} & \ul{.794} & \ul{.720} & {.932} & \ul{.798} & \ul{.777} & \ul{.950} & \ul{.841} \\ 
& MCLEA   &
  {.731} & {.943} & {.814} &{.789} &{.958} &{.854} & .814 & .967 & .873\\
 & \quad {\small w/  GMNM} & \ul{.759} & \ul{.964} & \ul{.840} & \ul{.806} & \ul{.974} & \ul{.871} & \ul{.837} & \ul{.980} & \ul{.893} \\ 
& {\small MEAformer}   
&  {.763} & {.963} & {.842} & {.811} & {.976} & {.874} & .844 & .980 &  .897\\
& \quad {\small w/  GMNM} & \ul{.779} & \ul{.968} & \ul{.847} & \ul{.817} & {.974} & \ul{.876} & \ul{.852} & \ul{.981} & \ul{.899} \\
& \ours~{\footnotesize (Ours)} & \textbf{.826} & \textbf{.976} & \textbf{.885} & \textbf{.852} & \textbf{.983} & \textbf{.904} & \textbf{.875} & \textbf{.987} & \textbf{.919} \\ 
% ------------------------------------ Open EA EN-FR-------------------------------------
\midrule
\parbox[t]{2mm}{\multirow{7}{*}{\rotatebox[origin=c]{90}{\small OpenEA$_{{\text{EN-FR}}}$}}} 
& EVA   
& {.605} & {.869} &{.700} & .619 & .870 & .710 & .848 & .973 & .896 \\
& \quad {\small w/  GMNM} & \ul{.606} & \ul{.870} & \ul{.701} & \ul{.621} & \ul{.874} & \ul{.713} & \ul{.856} & {.971} & \ul{.898} \\ 
& MCLEA   
& {.613} &{.889} & {.714} &{.702} &{.928} &{.785} & .893 & .983 & .928 \\
& \quad {\small w/  GMNM} & \ul{.625} & \ul{.902} & \ul{.726} & \ul{.707} & \ul{.934} & \ul{.790} & {.893} & {.983} & {.928} \\ 
& {\small MEAformer}   
&  {.660} & {.913} & {.751} & {.729} & {.947} & {.810}  & .895 & .984 & .930 \\
& \quad {\small w/  GMNM} & \ul{.666} & \ul{.916} & \ul{.755} & \ul{.741} & {.943} & \ul{.815} & \ul{.905} & {.984} & \ul{.937} \\ 
& \ours~{\footnotesize (Ours)} & \textbf{.692} & \textbf{.927} & \textbf{.778} & \textbf{.743} & \textbf{.945} & \textbf{.817} & \textbf{.907} & \textbf{.986} & \textbf{.939} \\ 
% ------------------------------------ Open EA EN-DE-------------------------------------
\midrule
\parbox[t]{2mm}{\multirow{7}{*}{\rotatebox[origin=c]{90}{\small OpenEA$_{{\text{EN-DE}}}$}}} 
& EVA   & {.776} &{.935} &{.833} & .784 &{.937} & .839 & .954 & .984 & .965 \\
& \quad {\small w/  GMNM} & \ul{.779} & \ul{.936} & \ul{.837} & \ul{.789} & \ul{.938} & \ul{.843} & \ul{.955} & {.984} & \ul{.966} \\ 
& MCLEA   
& {.766} & {.942} & {.829} &{.821} & {.956} &{.871} & .969 & .994 & .979 \\
& \quad {\small w/  GMNM} & \ul{.779} & \ul{.948} & \ul{.840} & \ul{.829} & \ul{.959} & \ul{.876} & \ul{.971} & \ul{.995} & \ul{.980} \\ 
& {\small MEAformer}   
&  {.803} & {.950} & {.854} & {.835} & {.958} & {.878} & .963 & .994 & .976 \\
& \quad {\small w/  GMNM} & \ul{.807} & {.949} & \ul{.856} & \ul{.841} & \ul{.961} & \ul{.882} & \ul{.975} & \ul{.995} & \ul{.982} \\ 
% & {UMAEA} {\footnotesize {\cite{chen2023rethinking}}}  
% &  {.771} & {.945} & {.833} & {.807} & {.956} & {.860} & .955 & .993 & .970 \\
& \ours~{\footnotesize (Ours)} & \textbf{.826} & \textbf{.962} & \textbf{.874} & \textbf{.859} & \textbf{.970} & \textbf{.899} & \textbf{.977} & \textbf{.998} & \textbf{.984} \\ 
% ------------------------------------ Open EA D-W-V1-------------------------------------
\midrule
\parbox[t]{2mm}{\multirow{7}{*}{\rotatebox[origin=c]{90}{\small OpenEA$_{{\text{D-W-V1}}}$}}} 
& EVA   & {.647} & {.856} & {.727} & .669 & .860 & .741 & .916 & .984 & .943 \\
& \quad {\small w/  GMNM} & \ul{.663} & \ul{.859} & \ul{.735} & \ul{.673} & \ul{.862} & \ul{.743} & \ul{.927} & \ul{.986} & \ul{.950} \\
& MCLEA   & {.686} &{.896} &{.766} &{.770} &{.941} &{.836} & .947 & .991 & .965 \\
& \quad {\small w/  GMNM} & \ul{.699} & \ul{.907} & \ul{.778} & \ul{.776} & \ul{.946} & \ul{.840} & \ul{.949} & {.991} & \ul{.966} \\
& {\small MEAformer}   
& {.718} & {.901} & {.787} & {.785} & {.934} & {.841} & .943 & .990 & .962 \\
& \quad {\small w/  GMNM} & \ul{.728} & {.901} & \ul{.793} & \ul{.803} & \ul{.942} & \ul{.855} & \ul{.956} & \ul{.991} & \ul{.970} \\
% & {UMAEA} {\footnotesize {\cite{chen2023rethinking}}}  
% &  {.668} & {.890} & {.750} & {.722} & {.908} & {.791} & .905 & .972 & .930 \\
& \ours~{\footnotesize (Ours)} & \textbf{.753} & \textbf{.930} & \textbf{.820} & \textbf{.808} & \textbf{.953} & \textbf{.864} & \textbf{.958} & \textbf{.993} & \textbf{.972} \\
% ------------------------------------ Open EA D-W-V2-------------------------------------
\midrule
\parbox[t]{2mm}{\multirow{7}{*}{\rotatebox[origin=c]{90}{\small OpenEA$_{{\text{D-W-V2}}}$}}} 
& EVA   & {.854} &{.980} &{.904} & .859 &{.983} & .908 & .925 & .996 & .951 \\
& \quad {\small w/  GMNM} & \ul{.866} & {.980} & \ul{.909} & \ul{.872} & {.981} & \ul{.913} & \ul{.948} & \ul{.997} & \ul{.969} \\
& MCLEA   &
  {.841} & {.984} & {.899} &{.877} & {.990} &{.923} & .971 & .998 & .983 \\
& \quad {\small w/  GMNM} & \ul{.845} & \ul{.987} & {.902} & \ul{.882} & \ul{.992} & \ul{.926} &\ul {.973} & \ul{.999} & \ul{.984} \\
& {\small MEAformer}  
&  {.886} & {.990} & {.926} & {.904} & {.992} & {.938}  & .965 & .999 &  .979 \\   
& \quad {\small w/  GMNM} & \ul{.902} & {.990} & \ul{.936} & \ul{.918} & \ul{.993} & \ul{.948} & \ul{.975} & {.999} & \ul{.985} \\
% & {UMAEA} {\footnotesize {\cite{chen2023rethinking}}} 
% &  {.846} & {.984} & {.897} & {.858} & {.986} & {.905} & .946 & .996 & .965 \\   
& \ours~{\footnotesize (Ours)} & \textbf{.904} & \textbf{.994} & \textbf{.939} & \textbf{.924} & \textbf{.994} & \textbf{.952} & \textbf{.980} & {.999} & \textbf{.988} \\
\bottomrule[0.8pt]
\end{NiceTabular}
}
\vspace{-5pt}
\label{tab:mmea-ap}
\end{table}